\documentclass[12pt]{article}

\usepackage{authblk}
\usepackage{geometry}
\usepackage{subcaption}
\usepackage{graphicx}
\usepackage{epstopdf}

\title{Universality of the $\pi^2/6$ Pathway in Avoiding Model Collapse}
\author{Apratim Dey\thanks{Corresponding Author. Email: \texttt{apd1995@stanford.edu}}}
\author{David Donoho\thanks{Email: \texttt{donoho@stanford.edu}}}
\affil{\textit{Department of Statistics, Stanford University}}

\date{October, 2024}

\usepackage{amsthm,amsmath,amsfonts,bm}

\newtheorem{theorem}{Theorem}[section]  % Theorems numbered within sections
\newtheorem{lemma}[theorem]{Lemma}      % Lemmas follow the numbering of Theorems
\usepackage{algorithm}
\usepackage{algpseudocode} % For algorithm writing
\usepackage{xcolor}
\newcommand{\aug}{\text{aug}}
\newcommand{\refer}{\texttt{ref}}

\newcommand{\hth}{\hat{\theta}}

\def\Secref#1{Section~\ref{#1}}
\def\eqref#1{equation~\ref{#1}}
\def\Eqref#1{Equation~\ref{#1}}
\def\1{\bm{1}}

\DeclareMathAlphabet{\mathsfit}{\encodingdefault}{\sfdefault}{m}{sl}
\SetMathAlphabet{\mathsfit}{bold}{\encodingdefault}{\sfdefault}{bx}{n}

\def\gB{{\mathcal{B}}}

\def\gD{{\mathcal{D}}}

\def\gF{{\mathcal{F}}}

\def\gN{{\mathcal{N}}}

\def\gV{{\mathcal{V}}}
\def\gW{{\mathcal{W}}}
\def\gX{{\mathcal{X}}}
\def\gY{{\mathcal{Y}}}
\def\gZ{{\mathcal{Z}}}

\def\sI{{\mathbb{I}}}

\def\sP{{\mathbb{P}}}
\def\sQ{{\mathbb{Q}}}
\def\sR{{\mathbb{R}}}

\newcommand{\E}{\mathbb{E}}

\newcommand{\R}{\mathbb{R}}

\newcommand{\Var}{\mathrm{Var}}

\newcommand{\Cov}{\mathrm{Cov}}
\DeclareMathOperator*{\argmin}{arg\,min}

\DeclareMathOperator{\Tr}{Tr}

\usepackage{hyperref}
\usepackage{url}
\usepackage{graphicx}
\usepackage{placeins}
\usepackage{float}
\usepackage{natbib}

\floatname{algorithm}{Workflow}

\begin{document}

\maketitle

\begin{abstract}

Researchers in empirical machine learning recently spotlighted 
their fears of so-called \textbf{Model Collapse}. They imagined a \textit{discard} workflow, where
an initial generative model is trained with real data, after which the real data are discarded, and subsequently, the model generates synthetic data on which a new model is trained. They came to the conclusion that models degenerate as model-fitting generations proceed. However, other researchers considered an \textit{augment} workflow, where the original real data continue to be used in each generation of training, {\it augmented} by 
synthetic data from models fit in all earlier generations. Empirical results comparing \textit{discard} and \textit{augment} workflows on canonical datasets and learning procedures confirmed the occurrence of model collapse under the \textit{discard} workflow and avoidance of model collapse under the \textit{augment} workflow. Under the \textit{augment} workflow, theoretical
evidence also confirmed avoidance in particular instances; specifically, \citet{gerstgrasser2024is} found that for classical Linear Regression, test risk at any later generation is bounded by a moderate multiple, viz. $\pi^2/6$ of the test risk of training with the original real data alone — no matter how many generations of training with synthetic data take place. Some commentators questioned the generality of theoretical conclusions based on the generative model assumed in \citet{gerstgrasser2024is}: could similar conclusions be reached for other task/model pairings? In this work, we demonstrate the \textit{universality} of the $\pi^2/6$ \textit{augment} risk bound, across a large family of canonical statistical models, offering key insights into exactly why collapse happens under the \textit{discard} workflow and is avoided under the \textit{augment} workflow. In the process, we provide a framework that is able to accommodate a large variety of workflows (beyond \textit{discard} and \textit{augment}) thereby enabling an experimenter to judge the comparative merits of multiple different workflows by simulating a simple Gaussian process.

\end{abstract}
\section{Introduction}\label{sec:intro}

Unbridled public access to text and image generative models (e.g. ChatGPT, Dall-E, Gemini, Claude, etc.) has brought about the concern that the Internet will soon be overwhelmed with artificially generated content. Since AI models are trained on public data, one imagines that future model trainings may soon be dominated by synthetic data. If synthetic data exhibit artifacts and distortions, these might compound across generations, leading to model degeneration,  a concern termed as \textit{Model Collapse} or \textit{Model Autophagy Disorder} \citep{shumailov2024ai,alemohammad2024selfconsuming,bertrand2024on,martinez2023combining,martinez2023towards,hataya2023will,dohmatob2024model,feng2024tale}. Researchers voicing this concern mostly 
considered a \textit{discard} workflow, where
one trains with real data at the initial generation, but in all later generations, 
one each time generates  synthetic data from the most recently trained previous model, 
and then trains on the resulting
synthetic data to produce the next generation of trained model. In this way,
the original real data are excluded from later generations of training. They showed that, 
for this workflow, trained model performance 
suffers \textit{unboundedly} across successive generations, 
tending towards eventual degradation. More recently, other researchers have considered an \textit{augment} workflow, where the original real data continue to be used in each generation of training, 
alongside synthetic data from models fitted in all previous generations. In specific instances, \citet{gerstgrasser2024is,marchi2024heat,seddik2024bad,kazdan2024collapse} have shown that model collapse is avoided by 
this workflow augmenting real data with synthetic data. In particular, \citet{gerstgrasser2024is} spotlight what we call a $\pi^2/6$ argument, whereby they show that under the \textit{augment} scenario, for the specific case of Linear Regression, the (excess) test risk is at most $\pi^2/6 (\approx 1.645)$ times that of the initial model trained only with real data.

Despite substantial and accumulating empirical evidence demonstrating avoidance of model collapse in the \textit{augment} scenario (e.g. \citet{gerstgrasser2024is} and follow-ups), published theoretical explanations of this phenomenon span only a small fraction of the cases where it has been observed empirically. The theoretical calculation leading to the $\pi^2/6$ bound in \citet{gerstgrasser2024is} is no doubt correct; but is seemingly specific to linear regression. A natural question emerges:

\paragraph{Question 1: }\textit{Is there a theoretical explanation for the reported empirical advantages of the \textit{augment} training workflow over the \text{discard} workflow, in particular  avoidance of model collapse in the  \textit{augment} workflow? }

\vspace{0.5cm}

\noindent
Of course, \textit{discard} and \textit{augment} are but only two of the potentially millions of workflows one may employ to iteratively fit models. In some sense, \textit{discard} and \textit{augment} are two pessimistic approaches in this regard, since they equally weigh all the (immediate past) synthetic data and all the (accumulated) real+synthetic data respectively, without accounting for data quality or data selection. Despite being pessimistic, that the \textit{augment} workflow enjoys a bounded error with the bound close to $1$, is certainly a pleasant surprise. Nonetheless, it is well known how daunting the challenge is in identifying real from synthetic data, and absent such a discriminatory tool, one would need to resort to these workflows.

Having said all this, it is indeed plausible that in the near future, high quality tools will emerge that will be increasingly able to differentiate between real and synthetic data. This translates to (confidently) downweighing potential synthetic data and upweighing potential real data for subsequent model training. With a bunch of data $Z_1,\cdots, Z_N$, one then optimizes a loss function $L(z;\theta)$ with weight $\omega_i$ associated with data point $Z_i$:
\begin{align}\label{eqn:general_weighting}
    \hat\theta \in\argmin_\theta \sum_{i=1}^N \omega_i L(Z_i;\theta)
\end{align}

Recently, towards potentially avoiding model collapse, researchers have considered a variety of approaches \citep{jain2024scaling,ferbach2024self,bertrand2023stability,alemohammad2024self,feng2024beyond,kazdan2024collapse}. Many of these approaches employ some kind of mixing or supervision to steer the models from deteriorating too fast as they iteratively get built, which implicitly correspond to \Eqref{eqn:general_weighting} with appropriately chosen (possibly data, model and generation dependent) weights $\omega_i$. 
To quote \citet{shumailov2024ai}: \color{violet}It is now clear that generative artificial intelligence (AI) such as large language models (LLMs) is here to stay and will substantially change the ecosystem of online text and images\color{black}. As a consequence, we expect a surge in research activity in exploring methods that attempt to ensure quality performance even when working with polluted data sources. However, it is well known that such mitigation techniques add extra computational burden to the training pipeline, and hence, it is highly desirable to be able to \textit{predict} which strategy (among a bunch of potential candidates) might be the best to employ, \textit{before} allocating resources to actually perform training. Thus, we arrive at another natural question:

\paragraph{Question 2:} \textit{Is it possible to theoretically understand the universe of possible collapse-mitigating approaches under a common unifying framework, and hence, given a pool of candidate approaches, provide informative advice on which one to use?}

\vspace{0.5cm}

As we show in this paper, the answer to both Questions 1 and 2 is \textbf{YES}.

\noindent

\section{Related works and our contributions}\label{sec:related_works}

\subsection{Prior works}

The original announcement of \citet{shumailov2023curse} (that is where the term \textbf{Model Collapse} was introduced) awakened the world into realizing that naively using only synthetic data for iterative model training leads to eventual, inevitable degradation. Since then, a flurry of intense research has ensued, driven both by empirical and theoretical scientists, who have confirmed the doomsday predictions through a wide variety of experiments and mathematical models \citep{alemohammad2024selfconsuming,bertrand2023stability,hataya2023will,martinez2023combining,martinez2023towards,dohmatob2024model,feng2024tale}. Popular media outlets have followed on, highlighting the perils of real data scarcity in an era where the Internet, the most important source of all human knowledge, gets polluted by indiscriminate amounts of poor quality, artificially generated data depositions.

Interestingly, much before the notion of model collapse became popular, in specific cases, mathematical scientists had already discussed the detrimental effects of iterative model fitting; \citet{mobahi2020self} have discussed the collapse of iteratively fit ridge estimators to $0$, and \cite{awan2020one} have warned how sampling synthetic data from a fitted statistical model usually results in inefficient estimators. In particular, \citet{awan2020one} point out that the asymptotic variance of the estimator obtained from synthetic data generated from a model fit on real data, is \textit{twice} that of the estimator obtained from real data itself. In the model collapse terminology, the work of \citet{awan2020one} is related to the second out of the potentially infinitely many model fitting generations, and can be extended to show that the variance of the estimator, and hence the (excess) test risk, scales as $G$ in generation $G$, thereby already predicting the scale at which the test risk explodes, characterized in subsequent works directly connected to model collapse \citep{shumailov2024ai,dohmatob2024model}.

Over time, researchers have proposed multiple potential solutions to mitigate the effects of model collapse, such as injecting real data at every model fitting generation, exploring regularization schemes, and some kind of supervision or feedback \citep{bertrand2023stability,martinez2023combining,martinez2023towards,jain2024scaling,alemohammad2024self,ferbach2024self,dohmatob2024model,feng2024beyond}. Such works usually assume that one \textit{knows} which data point is real and which is not (important for determining the right regularization parameter), or present methodologies that enable, in specific cases, steering away of models from getting worse over time. 

On the other hand, \citet{gerstgrasser2024is} have taken a completely different approach. They have considered an arguably much more realistic \textit{augment} workflow where the original real data source, despite being increasingly corrupted by synthetic data, is still used at every model fitting generation. Iteratively, the model at generation $G$ is fit using the original real data along with all intermediate synthetic data that have been generated by previous models, \textit{without} any discrimination between real and synthetic data. In this scheme, at generation $G$, the \textit{proportion} of original real data in the training corpus is $1/G$, and therefore dwindles to $0$ as more models are built. It is natural to imagine that such a workflow is doomed to failure as eventually the training corpus is awash with synthetic data; nevertheless, \citet{gerstgrasser2024is} present the pleasant surprise that it is possible to avoid model collapse in this workflow. Explicitly, by considering the case of simple linear regression, they establish that the (excess) test risk at any model-fitting generation is at most a decent multiple, $\pi^2/6 (\approx 1.645)$, of the (excess) test risk of the estimator fit on solely real data at the first generation - no matter how many generations are considered. Independently, \citet{marchi2024heat,seddik2024bad} also consider this \textit{augment} workflow and arrive at similar conclusions. In other words, these works highlight that a \textit{naive} experimenter may still get away without being doomed, as long as they use all the data, including both the original real data and the subsequent synthetic data, accumulated so far. \citet{gerstgrasser2024is,seddik2024bad,kazdan2024collapse} also provide a number of experiments to support their claims.

\subsection{Our contributions}

We establish a unified theoretical framework that not only addresses both Questions 1 and 2 (as mentioned in Section \ref{sec:intro}) simultaneously but also shows that there is no need to perform model-specific calculations. In the appropriate coordinates, for any given workflow, \textit{all} reasonable models exhibit the \textit{same} behavior, particularly when the dataset size is large. For example, this implies that viewed in the right coordinates, logistic regression behaves exactly as linear regression, thereby highlighting that linear regression is \textit{all} that is needed. At the core of this understanding is the behavior of a Gaussian process, which is straightforward to simulate and visualize. This result simplifies the landscape for researchers who may feel overwhelmed by the sheer variety of existing workflows, providing a clear basis for comparison. Theorem \ref{thm:main_thm} lays the foundation for this universality, with Lemmas \ref{lemma:discard} and \ref{lemma:augment}, along with the subsequent discussion, exploring the implications for particular workflows. We also provide examples to demonstrate how this principle applies to key metrics relevant to the machine learning community.

Our arguments are based on the notions of contiguity and local asymptotic normality, introduced by \citet{cam1960locally}. These have been traditionally used in theoretical statistics to perform power calculations of tests under alternative hypotheses \textit{close to the} null hypothesis, whereby tests can be compared with each other. It is indeed a pleasant surprise that such tools provide the most elegant answer \textit{at one go} in a completely different setup, of interest to modern machine learning, and present a clear understanding of the statistical behavior of self-consuming loops. 

\section{Background}\label{seq:background}

\subsection{Notations}

Unless otherwise stated, $\sP$ and $\E$ will stand for probability measure and expectation respectively. $\sP_0$ and $\E_0$ will, in particular, denote probability and expectation computed under the \textit{real} data generating distribution. For example, $\E_0(f(Z))$ will denote the expectation of $f(Z)$ where $Z$ comes from the real data generating distribution $\sP_0$. $\stackrel{p}{\to}$ will denote convergence in probability, while $\stackrel{d}{\to}$ will denote convergence in distribution. $o_\sP(1)$ will denote a quantity that converges to 0 in probability, under $\sP$. $Z$ will denote a generic data-point, often split into a feature $X$ and response $Y$, and hence we will often visualize $Z$ as a tuple $Z=(X,Y)$. For (possibly) random sequences $a_n,b_n$, $a_n\stackrel{n\to\infty}{\approx} b_n$ means $a_n=b_n(1+o_p(1))$ as $n\to\infty$. For a vector $x\in\sR^n$ and $r>0$, $\gB(x;r)$ will denote the ball $\{y\in\sR^n:\|y-x\|_2\leq r\}$.

\subsection{Preliminiaries} \label{sec:prelims}

\paragraph{The General Workflow.} Workflow \ref{alg:augment} describes a very general iterative model fitting procedure that subsumes all the procedures the literature so far has dealt with. We incrementally build up a chain of growing datasets $\gZ_1\subset \gZ_2\subset \cdots \subset \gZ_G \subset \cdots$. $\gZ_1$ contains the original pristine real data. Iteratively, at generation $g$, given $\gZ_g$, we create a dataset $\gD_{g+1}$ which has data of two types: data from a fixed / known model, denoted by $\gX_{g+1}$, and data from a \textit{learned} model (using, potentially, $\gX_{g+1}$ and parameter $\hat\theta_g^\aug$ estimated from $\gZ_g$), denoted by $\gY_{g+1}$. We combine the new data $\gD_{g+1}$ with all earlier data to get the enlarged dataset $\gZ_{g+1}=\gZ_g\cup\gD_{g+1}$. This workflow describes how the Internet evolves: the original real data still exists (since $\gZ_1\subset \gZ_g$ for all $g$) but gets increasingly contaminated with synthetic data $\gD_{g+1}$ produced at generation $g$. To the best of our knowledge, \citet{gerstgrasser2024is} first explicitly presented Workflow \ref{alg:augment}, albeit in the specific case of linear regression and with estimator $\hth_g$ computed by \textit{equally} weighing all the data points in $\gZ_g$. However, Step 3 in Workflow \ref{alg:augment} is general enough to cover any estimator one might envision.

\begin{algorithm*}
\caption{Iterative Model Training by Synthetic Data Augmentation}\label{alg:augment}
\begin{algorithmic}[1]
     \Require Positive integers $d_X,d_Y,d_\eta,d_\Theta$; parametric generative probability model $\{p(\cdot|\eta):\eta\in\R^{d_\eta}\}$ defined on $\R^{d_Y}$; function $\eta:\R^{d_X}\times \R^{d_\Theta}\to\R^{d_\eta}$.
     % ; and collection of known distributions $\{H_t\}_{t\geq 1}$ defined on $(\R^{d_X})^n$.
    \State Start with a dataset $\gZ_1=\{(X_{1,1},Y_{1,1}),\cdots (X_{1,n},Y_{1,n})\}$ where $X_{1,i}\in\R^{d_X}$ and $Y_{1,i}\in\R^{d_Y}$ for each $1\leq i\leq n$.
    \For{each generation $G\geq 1$}
        \State Estimate $\hat\theta_G$ from $\gZ_G$.
        \State Generate $\gX_{G+1}=\{X_{G+1,1},\cdots, X_{G+1,n}\}$.
        % with $(X_{g+1,1},\cdots, X_{g+1,n})\sim H_{g+1}$.
        \State Generate new $\gY_{G+1}=\{Y_{G+1,1},\cdots,Y_{G+1,n}\}$ with $Y_{G+1,i}\sim p(\cdot|\eta(X_{G+1,i},\hat\theta_G))$ independently for each $1\leq i\leq n$.
        \State Set $\gD_{G+1}=\{(X_{G+1,1},Y_{G+1,1}),\cdots, (X_{G+1,n},Y_{G+1,n})\}$.
        \State \textbf{Augment} the existing data corpus with the newly generated data: $\gZ_{G+1}=\gZ_G\cup\gD_{G+1}$.
    \EndFor
\end{algorithmic}
\end{algorithm*}

\paragraph{Exponential family model.} For clarity of exposition, we focus on a canonical setting, the exponential family of generative statistical models, which encompasses the vast majority of theoretical distributions used in model-building in statistics and machine learning.

Any exponential family of generative models can be described abstractly as follows.
It comprises probability densities of the form
\begin{align}\label{eqn:exp_family_model}
    p(y|\eta) &= \exp(\eta^\top T(y)-A(\eta))h(y),\hspace{0.2cm}y\in\R^{d_Y},\eta\in\Omega
\end{align}Here, $\eta$ is called the \textit{natural parameter}, $T:\R^{d_Y}\to\R^{d_\eta}$ is called the \textit{sufficient statistic}, $h(\cdot)$ is a function independent of the parameter $\eta$,
\begin{align}
    \Omega &= \{\eta\in\R^{d_\eta}:\int_{\R^{d_Y}}e^{\eta^\top T(y)}h(y)d\mu(y) <\infty\}
\end{align}and $A:\Omega\to\R$ is the \textit{log-partition function} defined by
\begin{align}
    \exp(A(\eta)) &= \int_{\R^{d_Y}}e^{\eta^\top T(y)}h(y)d\mu(y), \hspace{0.2cm}\eta\in\Omega,
\end{align}where $\mu$ is a $\sigma-$finite probability measure on $\R^{d_Y}$. Popular examples include normal, binomial, poisson, exponential, gamma, etc. distributions. It is well known \citep{lehmann1986testing} that exponential families allow convenient mathematical operations, making them particularly attractive for presenting the essential arguments without a barrage of assumptions or notations.

\paragraph{AAL estimators.} Given data $Z_1,\cdots, Z_N$, we consider weighted M-estimators $\hat\theta_N$:

\begin{align}\label{eqn:M_estim_argmin}
    \hat\theta_N \in \argmin_\theta \sum_{i=1}^N \omega_{N,i}L(Z_i;\theta) 
\end{align}where $L(\cdot;\cdot)$ is a loss function  and $\theta_0$ is the target parameter of interest.  The optimization in \eqref{eqn:M_estim_argmin} may in general yield multiple minimizers, but if the loss $L(z;\theta)$ is strictly convex in $\theta$ for each $z$, the minimizer is unique. Under standard regularity conditions on $L$, and when $Z_1,\cdots, Z_N$ are iid, $\hat\theta_N$ admits an asymptotic expansion of the form
\begin{align}
    \sqrt{N}(\hat\theta_N-\theta_0)&=\left(\dfrac{1}{N}\sum_{i=1}^N \omega_{N,i}\nabla^2_\theta L(Z_i;\theta)\right)^{-1}\left(\dfrac{1}{\sqrt{N}}\sum_{i=1}^N \omega_{N,i}\nabla_\theta L(Z_i;\theta)\right)+o_p(1)
\end{align}
Writing $A_N(\theta):=\sum_{i=1}^N \omega_{N,i}\nabla_\theta^2 L(Z_i;\theta)/N$, we can equivalently express the above as
\begin{align}\label{eqn:asymp_linear}
    \sqrt{N}(\hat\theta_N-\theta_0)=\dfrac{1}{\sqrt{N}}\sum_{i=1}^N \omega_{N,i}\psi(Z_i;\theta_0)+o_p(1)
\end{align}where $\psi(z;\theta)=A_N(\theta)^{-1}\nabla_\theta L(z;\theta)$. Usually, $A_N(\theta)$ concentrates around a deterministic quantity, yielding an \textit{asymptotically linear form} for the estimator $\hth_N$. Such an estimator $\hth_N$ satisfying \Eqref{eqn:asymp_linear} will therefore be called \textit{asymptotically approximately linear} (AAL). Common examples (under appropriate conditions on the model) include the (weighted versions of) sample mean, the sample median, estimators in linear regression, logistic regression, probit regression, quantile regression, etc.

As discussed previously, choosing different weights for data points is becoming increasingly important in modern machine learning, especially when one aims to emphasize \textit{high-quality} data to improve model performance or to mitigate the effects of distribution shift. Of course, $\omega_{N,i}=1$ for all $i$ covers the case of \textit{usual} unweighted M-estimators.

\subsection{Assumptions}

\paragraph{Assumption 1.} At each generation $G\geq 1$, the features $X_{G,1},\cdots, X_{G,n}$ are generated iid from a \textit{known} distribution $H$, free of $G$. Also assume that $H$ has finite moments of all orders.

\paragraph{Assumption 2.} 
At each generation $G\geq 1$, given feature $X_{G,i}$ and candidate parameter $\theta$, the response-generating distribution $p(\cdot|\eta(X_{G,i},\theta))$ comes from the exponential family model defined in \Eqref{eqn:exp_family_model} with natural parameter $\eta(X_{G,i},\theta)\equiv X_{G,i}\theta$ (identifying $\sR^{d_X}$ as $\sR^{d_\eta}\times \sR^{d_\Theta}$) and sufficient statistic $T(\cdot)$. That is,
\begin{align*}
    p(y|\eta(X_{G,i},\theta)) &= \exp(\theta^\top X_{G,i}^\top T(y) - A(X_{G,i}\theta))h(y)
\end{align*}We will also assume the following regularity condition. There exists $r>0$ such that for any (possibly random) $\tilde\theta\in \gB(\theta_0,r)$, and for any $X\in\R^{d_X}$,
\begin{align}
    \left|\dfrac{\partial^3A(\eta)}{\partial \eta_i\eta_j\eta_k}|_{\eta=X\tilde\theta}\right| \leq h(X), \hspace{0.2cm}\text{for all }1\leq i,j,k\leq d_\eta
\end{align}for a non-negative function $h$ which has finite moments (under $H$) of all orders.

\paragraph{Assumption 3.} For any $g, i\geq 1$, let $Z_{g,i}=(X_{g,i},Y_{g,i})$. At each generation $G\geq 1$, the estimator $\hth_{G}$ is AAL with
\begin{align}
    \sqrt{nG}(\hth_G-\theta_0) &= \dfrac{1}{\sqrt{nG}}\sum_{g=1}^G\sum_{i=1}^n \omega_{G,g,i}\psi(Z_i;\theta) + o_{\sP_0}(1)
\end{align}where $\omega_{G,g,i}$ is a (possibly random) weight associated with $Z_{g,i}$ but is independent of $Z_{g,i}$, $\psi(z;\theta)= A_{n,G}(\theta)^{-1}\nabla_\theta L(z;\theta)$ for a loss function $L(\cdot;\cdot)$,
\begin{align}
    A_{n,G}(\theta) &= \dfrac{1}{nG}\sum_{g=1}^G\sum_{i=1}^n \omega_{G,g,i}\nabla^2_\theta L(Z_{g,i};\theta)
\end{align}and $\E_0[\nabla_\theta L(Z;\theta_0)]=0$.

\paragraph{Comments on assumptions.} These assumptions are framed primarily to maximize efficiency of exposition. Our goal here is \textbf{not} to present the most general conditions under which the stated result holds; rather, we hope to introduce a broadly applicable, orthodox theoretical statistics framework
within which one clear consistent conclusion emerges about the two different workflows under consideration. 
We also maintain that a researcher may use this framework to understand the performance of some other workflow for learning from synthetic and real data, or some other family of estimators.

Assumption 1 underlines that we \textit{do not} learn the feature distribution, and that they are generated from a known distribution $H$. This semi-parametric approach is adopted to primarily \textit{separate} the training and non-training components; if we do learn the feature distribution per generation, we can disregard $X$ altogether and identify a data point $Z$ with the response $Y$. We also assume the distribution of the features, $H$, does not change from generation to generation. This again is chosen for convenience; one may easily accommodate the case where it changes from generation to generation by straightforward adaptation of the proof. Further, it is also possible to weaken Assumption 2 to accommodate the case when the features are recycled per iteration, although to keep the results the cleanest, we do not discuss this.

Assumption 2 imposes a generalized linear exponential family model on the response $y$ conditional on the feature $X$. Most modern architectures employ a generalized linear model at their top layer (e.g. the softmax layer for image classification or language models corresponds to multinomial logistic distribution). Note that the raw data might be of the form $(F,Y)$ where $F$ is a \textit{raw} feature vector. We are not imposing the restriction that the natural parameter be linear in the raw feature $F$, rather linear in \textit{some} transformation of $F$, namely, $X$, which, by extension, we still call feature. For pre-trained image classification or language models, during forward pass, the raw features (pixels or tokens) are converted into learned features, and a generalized linear model is fit on these learned features obtained in the penultimate layer, which we identify as $X$ (the label being identified as $Y$). The regularity condition assumed is standard in asymptotic theory, see \citet{van2000asymptotic} for example. It allows control of remainder terms when $A(\cdot)$ is Taylor expanded.

Finally, coming to Assumption 3, as discussed in \Secref{sec:prelims}, a large collection of M-estimators used in standard machine learning models are actually AAL. Furthermore, assuming an AAL estimator makes the key ingredients of our proof method visibly transparent.
\section{Comparison of Experiments and Statistical Efficiency}\label{sec:comparison_experiments}

Classical statistical decision theory, formulated in the 1920's-1950's, delivered a theoretically airtight way to make statements
of the form:``this estimator can achieve the same results as this other estimator, using only a fraction of $f$ as much data". 
The key heuristics go back to Sir RA Fisher, but today's final formalization is credited to Abraham Wald and Jacob Wolfowitz, Erich Lehmann and, in its most advanced conceptualization, Lucien Le Cam and Jaroslav Hajek; see the discussion in \citet{lehmann1986testing}.
Suppose that estimator $\hth_1$ achieves mean-squared error $MSE(\hth_1, n_1)$ on a sample of size $n_1$ and estimator $\hth_2$ achieves $MSE(\hth_2, n_2)$ on a sample of size $n_2$. Consider `proportionally growing' sequences where $n_1 /n_2 \rightarrow f $ as each $n_i \rightarrow \infty$, for some $f \in (0,\infty)$. Suppose that:
\[
     \frac{MSE(\hth_2,n_2)}{MSE(\hth_1,n_1)} \rightarrow 1, \qquad n_2 \rightarrow \infty, \qquad n_1 \sim f \cdot n_2 \rightarrow \infty.
\]
Then we say that the {\it asymptotic relative efficiency of $\hth_2$ relative to $\hth_1$} is $f \times 100$ percent. 
% Intuitively, if $f = 1/2$, then $\hth_1$ needs `half as much data' as $\hth_2$ to achieve the same quality of estimate. We write this so: $ARE(\hth_2;\hth_1) = f$.

% Today, there is much discussion about how data quality matters over quantity and how scaling laws can be significantly improved with better quality data. Judging the quality of a dataset, or of a procedure for that matter, is by no means an easy task. Multiple researchers may propose multiple experiments, and it is crucial to have a formalism that can offer a seamless comparison among those. For example, multiple researchers have studied multiple different methods of iterative model training with real and synthetic data - some consider purely synthetic, some inject fresh real data per iteration, some (implicitly) overweigh the real data during model training, etc. \citationsneeded Usually the performance of such procedures is analyzed using crude bounds and concentration inequalities, which, despite offering a certain degree of insight, do not provide the most precise information. Multiple estimators may have the same \textit{rate} of concentration around the desired quantities, but their real performance depends quite substantially on the constants. Fortunately, this problem has been addressed classically in statistics through the notion of \textit{asymptotic relative efficiency} (ARE). 
Consider the particular case where the estimators are root-$n$ consistent and asymptotically normal, i.e. obey:
\begin{align}\label{eqn:ARE_normality}
    \sqrt{n}(\hth_i-\theta_0) \stackrel{\text{approx,} n\to\infty}{\sim} \gN(0, V_i),\hspace{0.2cm}i=1,2 .
\end{align}
In such cases $MSE(\hth_i,n_i) \sim V_i/n_i$,
so the required sample size ratio for equal performance
$MSE(\hth_1,n_1) ~ MSE(\hth_2,n_2)$ obeys:
$n_1/n_2 \rightarrow  f = V_2/V_1$.
The ARE of $\hth_2$ relative to $\hth_1$ is, therefore, simply the 
\textit{inverse} of the ratio of the asymptotic variances, viz.
\begin{align}\label{eqn:ARE_comparison}
    ARE(\hth_2;\hth_1) &= V_1/V_2 .
\end{align}
If $V_1 > V_2$ we conclude that $\hth_2$ is at least as good as $\hth_1$,
and can also consider $ARE(\hth_1; \hth_2)$ reversing the order of arguments. 
Note that a wide collection of parameter estimation procedures studied in machine learning and statistics are based on \textit{empirical risk minimization} (ERM) and usually they admit a limiting Gaussian approximation as in (\ref{eqn:ARE_normality}).

% In the above comparisons we may choose various different estimators to play the roles of  $\hth_1$ and $\hth_2$ and achieve different types of insights. In this work our main results consider comparisons between estimators trained without any synthetic data, versus the {\it same estimators} trained with some synthetic data. 
% Typically, in our comparisons we consider a specific fixed estimation rule and model generative model, By ruke we mean a prescription for generating estimates for any given dataset, including datasets of various sizes. 
The role of $\hth_1$ in our comparisons is
played by the traditional way of applying our  considered estimation rule \textit{applied only to the $n$ real data}.
The role of $\hth_{2}$ is played by that {\it same estimation rule}, however the rule applied to a nontraditional dataset, which may be of the same size or a larger size. 
% Namely, as we study either the discard or augment workflows, we apply the considered rule to the
% $n$ data produced at the $G$-th generation
% of synthetic data (discard workflow);
% or to all data real and synthetic up to and including the
% $G$-th generation (augment workflow).
In the \textit{discard} and \textit{augment} cases, we denote the resulting estimates by  
$\hth_{G}^{dis}$ (discard) 
and $\hth_{G}^{aug}$ (augment).
% ; and depending on
% choice, the role of $\hat{\theta}_2$ 
% in our risk comparisons could
% be played by either  $\hth_{G}^{dis}$
% or $\hth_{G}^{aug}$. 

Our main results consider a  wide variety of canonical statistical models,
and we show that, under the  \textit{augment} workflow, the asymptotic relative efficiency
is always bounded below by $\mathbf{60\%}$, no matter how many synthetic data generations take place. More precisely: at {\it any} generation $G$,  
\begin{align}
     ARE(\hth^{aug}_{G};\hth_1) \geq 6/\pi^2 > 60\%, \qquad \forall G \geq 1;
\end{align}
in contrast,  under the \textit{discard} workflow, 
efficiency degrades with increasing $G$:
\begin{align}
    ARE(\hth^{dis}_{G};\hth_1)  \rightarrow 0, \qquad G \rightarrow \infty.  
\end{align}

% Since in each case the baseline estimator $\hth_0$ is the same, \textit{augment} over \textit{discard} increasingly risk-dominates  as the number of generations increases. 
This clearly establishes the superiority of the \textit{augment} scheme. 

Our framework for risk comparison goes significantly beyond the two specific schemes studied in this paper, and can accommodate a spectrum of schemes that an experimenter may come up with: for example, unequal and possibly random cross-generation data weighting. We discuss this later.

Furthermore, while the Gaussian approximation in \Eqref{eqn:ARE_normality} happens at the level of the parameter, one can derive from it formal quantitative expressions for other metrics that might be of interest to the problem at hand. If $m(\theta)$ is such a twice-differentiable metric, e.g. test loss (evaluated at a candidate $\theta$), then under standard conditions on $m$, we can apply the delta method to conclude
\begin{align}
    \sqrt{n}(m(\hth_i)-m(\theta_0)) \stackrel{\text{approx,}n\to\infty}{\sim}\gN(0, V_im''(\theta_0))
\end{align}Thus, the asymptotic variance of $m(\hth_i)$ (in estimating $m(\theta_0)$) is $V_im''(\theta_0)$, which implies that
\begin{align}
    ARE(m(\hth_2);m(\hth_1)) &= V_1/V_2 = ARE(\hth_2;\hth_1).
\end{align}In particular, 
\begin{align}
    ARE(m(\hth^{aug});m(\hth_1)) \geq 6/\pi^2, \qquad \forall G\geq 1;\\
    ARE(m(\hth^{dis});m(\hth_1)) \to 0, \qquad G\to\infty
\end{align}
As a result, quantitative estimates obtained at the parameter level seamlessly \textit{transfer} to other metrics of practical interest. Later, we will provide examples for specific metrics of importance to the machine learning community to illustrate the power of our result.

\section{Main results}\label{sec:main_results}

\subsection{Main Theorem}

The main result establishes an asymptotic equivalence between \textit{any} sequence of model-fitting iterations and a Gaussian sequential experiment. In the large sample limit, the statistical properties of the model-fitting procedure align with those of a Gaussian process. Consequently, the specific complexities of the model-fitting process become irrelevant, allowing one to focus \textit{solely} on the Gaussian process. This equivalence offers a powerful and unifying perspective, enabling researchers to analyze the Gaussian process alone and thus make confident predictions about their experiment.

\begin{theorem}\label{thm:main_thm}
    Make Assumptions 1, 2 and 3. Define, for any $G\geq 1$,
    \begin{align}
        W_{n,T}(G) &= \dfrac{1}{\sqrt{n}}\sum_{i=1}^nX_{G,i}^\top(T(Y_{G,i})-\nabla A(X_{G,i}\theta_0)),\nonumber\\
        W_{n,\Theta}(G) &= \sqrt{n}(\hth_G-\theta_0)
    \end{align}Consider a reference distribution $\sP^\refer$ that assumes that at each generation $G\geq 1$, the data points (accumulated so far) in $\gZ_G$ are iid. That is, under $\sP^\refer$, for every $G\geq 1$ and $1\leq i\leq n$, $X_{G,i}$ are iid from $H$ and
    \begin{align*}
        Y_{G,i}|X_{G,i}\sim \exp(\theta_0^\top X_{G,i}^\top T(Y_{G,i})- A(X_{G,i}\theta_0))h(Y_{G,i}) 
    \end{align*}Then, the following hold.
    \begin{enumerate}
        \item[(a)]      Under $\sP^\refer$, the following distributional convergence holds as $n\to\infty$:
    \begin{align}
        (W_{n,T}(g), W_{n,\Theta}(g)_{g=1}^G
        \stackrel{d}{\to} (W_T^\refer(g),W^\refer_\Theta(g))_{g=1}^G
    \end{align}where $(W_T^\refer(g),W_\Theta^\refer(g))_{g=1}^G$ is a collection of jointly Gaussian mean zero variables.
    \item[(b)] As $n\to\infty$, under the actual sequential data generating mechanism presented in Workflow \ref{alg:augment},
    \begin{align}
        (W_{n,T}(g), W_{n,\Theta}(g)_{g=1}^G
        \stackrel{d}{\to} (W_T(g),W_\Theta(g))_{g=1}^G
    \end{align}where $(W_T(g),W_\Theta(g))_{g=1}^G$ denotes a sequential Gaussian process defined by the following rules:
    \begin{itemize}
        \item Generate $$W_T(g) = W_T^\refer(g) + \E_0[X^\top \nabla^2 A(X\theta_0)X]W_\Theta(g-1)$$with $W_T^\refer(g)$ independent of $W_\Theta(g-1),W_T(g-1),\cdots, W_T(1)$
        \item Given $\{W_T(g),W_\Theta(g-1),\cdots, W_T(1)\}$, generate $W_\Theta(g)$ from the reference conditional distribution:
        \begin{align}
            W_\Theta(g) &\sim \sP^\refer\left(\cdot\mid W_T^\refer(g)=W_T(g),W_\Theta^\refer(g-1)=W_\Theta(g-1),\cdots, W_T^\refer(1) = W_T(1)\right)
        \end{align}
        % W_\theta(g) &\sim W_\theta^\refer(g)\mid W_T^\refer(g),W_\theta^\refer(g-1),\cdots, W_T^\refer(g)
    \end{itemize}
    \end{enumerate}
\end{theorem}

\paragraph{Remark.}We make a few comments on Theorem \ref{thm:main_thm}.
We refer to Section \ref{sec:proof_structure} for insights into the proof. 
\begin{enumerate}
    \item The first part is a simple application of the Central Limit Theorem, thanks to the reference distribution $\sP^\refer$ that ensures $(X_{g,i},Y_{g,i})$ are all iid.
    \item The second part crucially hinges on the distribution derived in the first part. In words, the limiting Gaussian process evolves in the following way. At generation $g$, $W_T(g)$ develops a (random) mean which is a (possibly matricial) multiple of $W_\Theta(g-1)$. This marks the deviation from $W_T^\refer(g)$.  However, $W_\Theta(g)$ is (still) generated following the \textit{same} conditional distribution (given the past) as under the reference distribution $\sP^\refer$.
    
    \item The conditional distributions in $\sP^\refer$ are conditional Gaussian distributions. Hence, they can be computed explicitly recalling the formula for the conditional Gaussian distribution. 
    % In addition, this result and its theoretical implications encompass a large variety of models studied in the literature on model collapse.
    \item The development of the random mean, $\E_0[X^\top \nabla^2 A(X\theta_0)X]W_\Theta(g-1)$, adds extra noise to $W_T(g)$, thereby enlarging its variance compared to $W_T^\refer(g)$.
\end{enumerate}

\subsection{Implication on estimation}
We now exhibit the power of Theorem \ref{thm:main_thm} in theoretically analyzing two commonly studied workflows in the literature on model collapse. Several researchers have previously focused on the \textit{discard} workflow and highlighted collapse in specific theoretical models \citep{shumailov2024ai,alemohammad2024selfconsuming,dohmatob2024model,bertrand2024on}. By choosing a specific sequence of weights that correspond to using only the immediate past synthetic data and discarding the previous history, we establish that collapse is bound to happen in such a case for a wide variety of models, going significantly beyond the scope of the previous literature. We are not aware of any result at this level of generality.
\begin{lemma}[\textit{Discard} workflow]\label{lemma:discard}
    Consider the sequence of weights 
    \begin{align}
        \omega_{G,g,i} = \begin{cases}
            1, & g=G, 1\leq i\leq n,\\
            0, & \text{otherwise}
        \end{cases}
    \end{align}Then, $\Var(W_\Theta(G))=G\times \Var(W_\Theta(1))$. As a result, $$ARE(\hat\theta_G^{dis};\hat\theta_1)=1/G\stackrel{G\to\infty}{\to}0$$
\end{lemma}

Next, we study the \textit{augment} workflow introduced in \citet{gerstgrasser2024is}, who also spotlight the $\pi^2/6$ bound discussed previously, albeit in the case of linear regression. Again one merely needs to choose the weights appropriately; in this case, the weights are all $1$, corresponding to equal importance given to all the data points accumulated thus far. Theoretical results exist only in specific cases so far \citep{gerstgrasser2024is,marchi2024heat,seddik2024bad,kazdan2024collapse}. The following Lemma \ref{lemma:augment} proves the essential universality of the $\pi^2/6$ bound.

\begin{lemma}[\textit{Augment} workflow]\label{lemma:augment}
    Consider the sequence of weights 
    \begin{align}
        \omega_{G,g,i} = \begin{cases}
            1, & 1\leq g\leq G, 1\leq i\leq n,\\
            0, & \text{otherwise}
        \end{cases}
    \end{align}Then, $\Var(W_\Theta(G))= \Var(W_\Theta(1))\times \left(\sum_{g=1}^G1/g^2\right)$. As a result, $$ARE(\hat\theta_G^{aug};\hat\theta_1)=\left(\sum_{g=1}^G 1/g^2\right)^{-1}\geq 6/\pi^2 > 60\%$$
\end{lemma}

\subsection{Implication on prediction}

Lemmas \ref{lemma:discard} and \ref{lemma:augment} describe the asymptotic MSE of the estimators $\hth_G^{dis}$ and $\hth^{aug}_G$ respectively. Can we also understand the behavior of test losses commonly used in machine learning tasks, as model fitting generations progress? To answer this, we consider a particular loss, the cross entropy (CE) loss, evaluated on a test point $Z=(X,Y)$, when $Z$ actually comes from $\sP_0$ but we fit the distribution $p(\cdot|X\hth)$ to the conditional distribution of $Y|X$. The evaluation of CE loss is closely related to the evaluation of the loglikelihood ratio $\log(p(Y|X\hth)/p(Y|X\theta_0))$:
\begin{align}
    CE(\hth) &= -\E_0\left(\log\left(\dfrac{p(Y|X\hth)}{p(Y|X\theta_0)}\right)\mid\hat\theta\right) - \E_0(\log(p(X,Y)))
\end{align}Note that the second term on the right is a constant. In fact, $CE(\hth)+\E_0(\log(p(X,Y)))$ is nothing but the (test) Kullback Leibler divergence between the true and fitted distributions, $D_{KL}(p(Y|X\hth)\|p(Y|X\theta_0))$, to be abbreviated as $D_{KL}(\hth\|\theta_0)$. Thus we can write
\begin{align}
    CE(\hth) &= D_{KL}(\hth\|\theta_0) - \E_0(\log(p(X,Y)))
\end{align}
In what follows, we compare the (expected) Kullback-Leibler test losses obtained from the \textit{discard} and \textit{augment} workflows, with that obtained by using estimator $\hth_1$ fitted using solely real data from generation $1$.

\begin{lemma}\label{lemma:discard_test}
    Make Assumptions 1, 2 and 3. Under the \textit{discard} workflow defined in Lemma \ref{lemma:discard}, for any model fitting generation $G\geq 1$, formally
    \begin{align}
        \dfrac{\E D_{KL}(\hth_G^{dis}\|\theta_0)}{\E D_{KL}(\hth_1\|\theta_0)} \stackrel{n\to\infty}{\approx} G
    \end{align}
\end{lemma}

\begin{lemma}\label{lemma:augment_test}
        Make Assumptions 1, 2 and 3. Under the \textit{augment} workflow defined in Lemma \ref{lemma:augment}, for any model fitting generation $G\geq 1$, formally
    \begin{align}
        \dfrac{\E D_{KL}(\hth_G^{aug}\|\theta_0)}{\E D_{KL}(\hth_1\|\theta_0)} \stackrel{n\to\infty}{\approx} \left(\sum_{g=1}^G \dfrac{1}{g^2}\right)\leq \dfrac{\pi^2}{6}
    \end{align}
\end{lemma}

Lemmas \ref{lemma:discard_test} and \ref{lemma:discard_test} quantify how far the fitted distributions get from the true distribution, highlighting linear rate of divergence under \textit{discard} and bounded (the bound being the same $\pi^2/6$) divergence under \textit{augment}. The derivations are provided in Section \ref{sec:appdx_predictive_loss}. For regression problems, the least squares loss is more relevant. We will discuss it in Section \ref{sec:examples}.

\subsection{Implication on computation and comparison}

Multiple researchers may propose multiple different workflows (that is, multiple different data weighing schemes). Some of them may be as simple as equally weighing or random subsampling, whereas others may be more complicated (e.g. applying data selection strategies to cherry-pick high quality data for model building). For only a handful of such procedures can one hope to obtain precise and aesthetic asymptotic characterization of the MSE of the estimator. Lemmas \ref{lemma:discard} and \ref{lemma:augment} describe the MSE of estimators under the discard and augment workflows respectively. However, an example of a workflow where getting an explicit expression for the MSE seems to be challenging, is the so-called \textit{augment-subsample} workflow. In this new workflow, at generation $G$, we randomly choose $n$ out of $nG$ data points in $\gZ_G$ to fit the model, and the weights reflect this selection: $(\omega_{G,g,i})_{g,i}$ is a uniformly random draw from the set of $nG$-dimensional binary $(0/1)$ tuples each having exactly $n$ $1$'s. This workflow is attractive as it seems to \textit{combine} the best of \textit{discard} and \textit{augment} - each subsequent model only trains on $n$ data points selected uniformly at random from the entire collection of $nG$ data points (containing both real + synthetic data).

\begin{figure*}[h]
% \vspace{.3in}
\centering
    \begin{subfigure}[b]{0.4\textwidth}
        \centering
        \includegraphics[width=\textwidth]{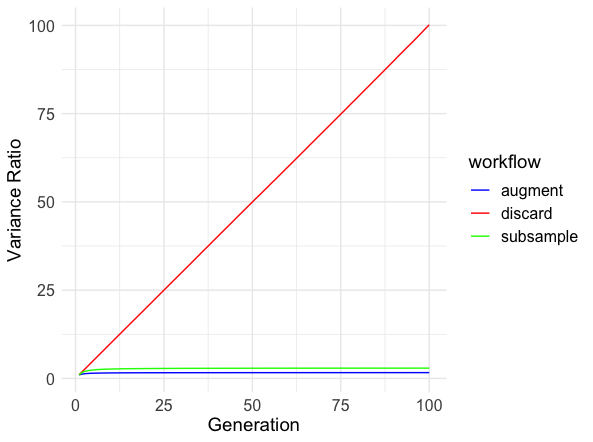}
    \end{subfigure}
    \begin{subfigure}[b]{0.4\textwidth}
        \centering
        \includegraphics[width=\textwidth]{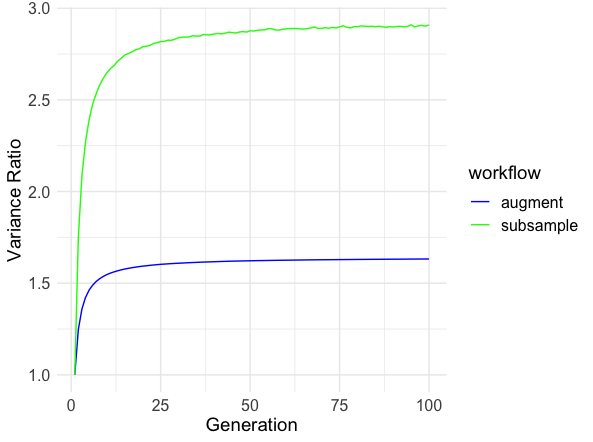}
    \end{subfigure}
% \vspace{.3in}
\caption{Plots showing ratio of variances of limit Gaussian variables across generations for three different workflows. The plot on the right is a zoomed-in version of the plot on the left (excluding the curve corresponding to \textit{discard}). It is clear that only \textit{discard} workflow exhibits exploding variance ratio, linearly as predicted by theory. The \textit{augment} workflow quickly concentrates around $\pi^2/6$. The \textit{augment-subsample} workflow also seems to plateau albeit at a higher value.}
\label{fig:gaussian_limits}
\end{figure*}

However, it is remarkably simple to simulate the limit Gaussian process, as one only needs to know the distributions under $\sP^\refer$. In Figure \ref{fig:gaussian_limits}, we perform a simulation study comparing the \textit{discard}, \textit{augment} and \textit{augment-subsample} workflows. We simulate a million trajectories of the limit Gaussian process for a Normal mean estimation problem where the estimator used is the sample median. We plot the ratio of the variances at 100 generations to the variance at generation $1$ of the Gaussian limits. Very clearly, the curve corresponding to \textit{discard} grows linearly with generation count $G$, as predicted by our theory. The curve corresponding to \textit{augment} converges to $\pi^2/6$, while the curve corresponding to \textit{augment-subsample} also seems to stabilize at a higher value (although whether it converges or diverges slowly is unclear). This lends firm support to the intuitive ordering: 
\begin{align*}
    ARE(\hat\theta_G^{aug},\hat\theta_1) > ARE(\hat\theta_G^{sub},\hat\theta_1) > ARE(\hat\theta_G^{dis},\hat{\theta_1})
\end{align*}

\section{Examples}\label{sec:examples}

We now provide some explicit examples of common models that exhibit the power of our main result. 

\subsection{Linear Models}

Linear models are perhaps the most ubiquitous models in statistics and machine learning. The linear model imposes a linear relation $y=x^\top\beta + e$ where $e\sim \gN(0,\sigma^2)$. This corresponds to an exponential family model with $\theta=(\beta/\sigma^2,1/\sigma^2)$ and sufficient statistic $T(y)=(y,y^2)$. The feature $X$ (constructed from the raw features $x$) such that $\eta(x,\theta)=X\theta$ is given by 
\begin{align}
    X = \begin{bmatrix}
        x^\top & 0\\
        0 & -1/2
    \end{bmatrix}
\end{align}
\citet{shumailov2024ai}, \citet{dohmatob2024model} and \citet{gerstgrasser2024is} have considered linear models. When $\sigma$ is unknown and is to be estimated, exact calculations, despite being still doable in the \textit{discard} case, become quite cumbersome in the \textit{augment} case.

As we shall see now, we can avoid all the messy calculations and deliver all the results at once. Previously, scientists have computed explicitly the means and variances of $\hat\beta$ and $\hat\sigma^2$ as generations progress, and the main point we would like to highlight is that the most convenient approach is to not deal with $(\beta,\sigma^2)$ but rather with the natural parameter $\theta$. In the case when one has data $(x_i,y_i)_{i=1}^N$ following the linear model, the most natural way to estimate $\theta_0=(\beta_0/\sigma_0^2,1/\sigma_0^2)$ is to use the maximum likelihood estimator 
   $ \hat\theta = (\hat\beta/\hat\sigma^2,1/\hat\sigma^2)$ given by
   \begin{align}
       \hat\beta &= \argmin_\beta \sum_{i=1}^N (y_i - x_i^\top\beta)^2\\
       \hat\sigma^2 &= \dfrac{1}{N}\sum_{i=1}^N (y_i - x_i^\top\hat\beta)^2
   \end{align}
 The MLE $\hat\theta$, of course, is an M-estimator (as it minimizes the negative log-likelihood), and thus, under suitable conditions on the weights, our assumptions hold. As a consequence, we immediately obtain a result of the following form:
\begin{align*}
    \sqrt{n}\begin{pmatrix}
        \dfrac{\hat\beta_G}{\hat\sigma^2_G}-\dfrac{\beta_0}{\sigma_0^2}\\
        \dfrac{1}{\hat\sigma^2_G}-\dfrac{1}{\sigma_0^2}
    \end{pmatrix}\stackrel{n\to\infty}{\to}\gN(0, V_G)
\end{align*}where $V_G$ is the asymptotic covariance matrix satisfying $V_G = G\times V_1$ for \textit{discard} and $V_G = (\sum_{g=1}^G 1/g^2)\times V_1$ for \textit{augment}. In particular, in the limit $n\to\infty$,
\begin{align*}
    \hat\sigma_G^2 \approx \dfrac{1}{1/\sigma_0^2+s_G\gW/\sqrt{n}}
\end{align*}for a (non-degenerate) gaussian $\gW$, where $s_G=\sqrt{G}$ for \textit{discard} and $s_G=\sqrt{\sum_{g=1}^G 1/g^2}$ for \textit{augment}. For \textit{discard} workflow, this shows that $\hat\sigma_G^2\to 0$, retrieving a result that previous authors have obtained \citep{alemohammad2024selfconsuming} by direct calculation (that too, in the case when $\beta_0=0$). On the other hand, as $s_G<\sqrt{\pi^2/6}$ no matter how large $G$ is, the \textit{augment} workflow ensures that $\hat\sigma^2_G$ does not decay to $0$ and remains as a positive random variable even in the limit of infinite iterations.

This reinforces the fact that tracking the limiting distribution empowers one to make conclusions about all possible aspects of the linear models completely avoiding any cumbersome calculation.

\subsection{Logistic Models}

The logistic is another highly popular exponential family model used in the (top) softmax layer for modern neural nets to fit binary or multinomial responses. For the binary case, for example, it corresponds to the natural parameter $\eta(X,\theta)=X\theta$ with $X=x^\top$ and sufficient statistic $T(y)=y\in\{0,1\}$. Direct calculations are difficult in this model. The most popular method to estimate the parameter $\theta$ is by maximum likelihood estimation, which is an M-estimator. As a result, the same $\pi^2/6$ bound will hold in the \textit{augment} workflow according to Lemma 4.3. This immediately extends the linear regression example considered in \citet{gerstgrasser2024is} to logistic regression for \textit{augment} workflow, thereby presenting the $\pi^2/6$ bound. Of course, it also immediately tells us that the \textit{discard} workflow behaves identically as well - the variance of $\hat\theta_G$ explodes linearly in logistic regression too, extending the conclusion derived in linear regression by \citet{dohmatob2024model}. Thus, key insights from the linear regression settings considered by previous authors transfer immediately to an apparently more complicated model as logistic regression \textit{without any further calculation}.
\section{Experiments}\label{sec:experiments}

\begin{figure*}[h]
    \centering
    \begin{subfigure}[b]{0.3\textwidth}
        \centering
        \includegraphics[width=\textwidth]{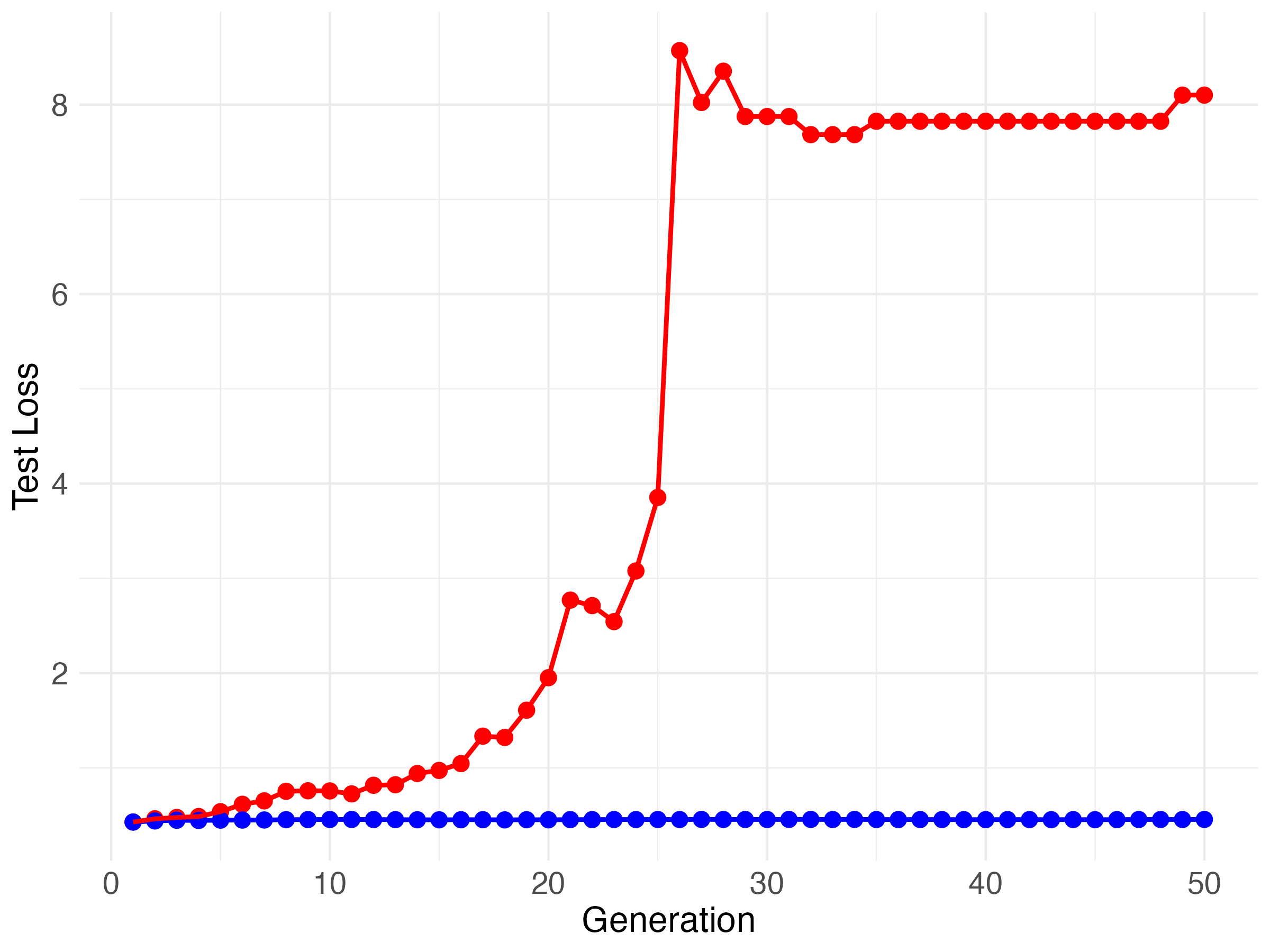}
        \subcaption{Diabetes}
        \label{fig:diabetes}
    \end{subfigure}
    \begin{subfigure}[b]{0.3\textwidth}
        \centering
        \includegraphics[width=\textwidth]{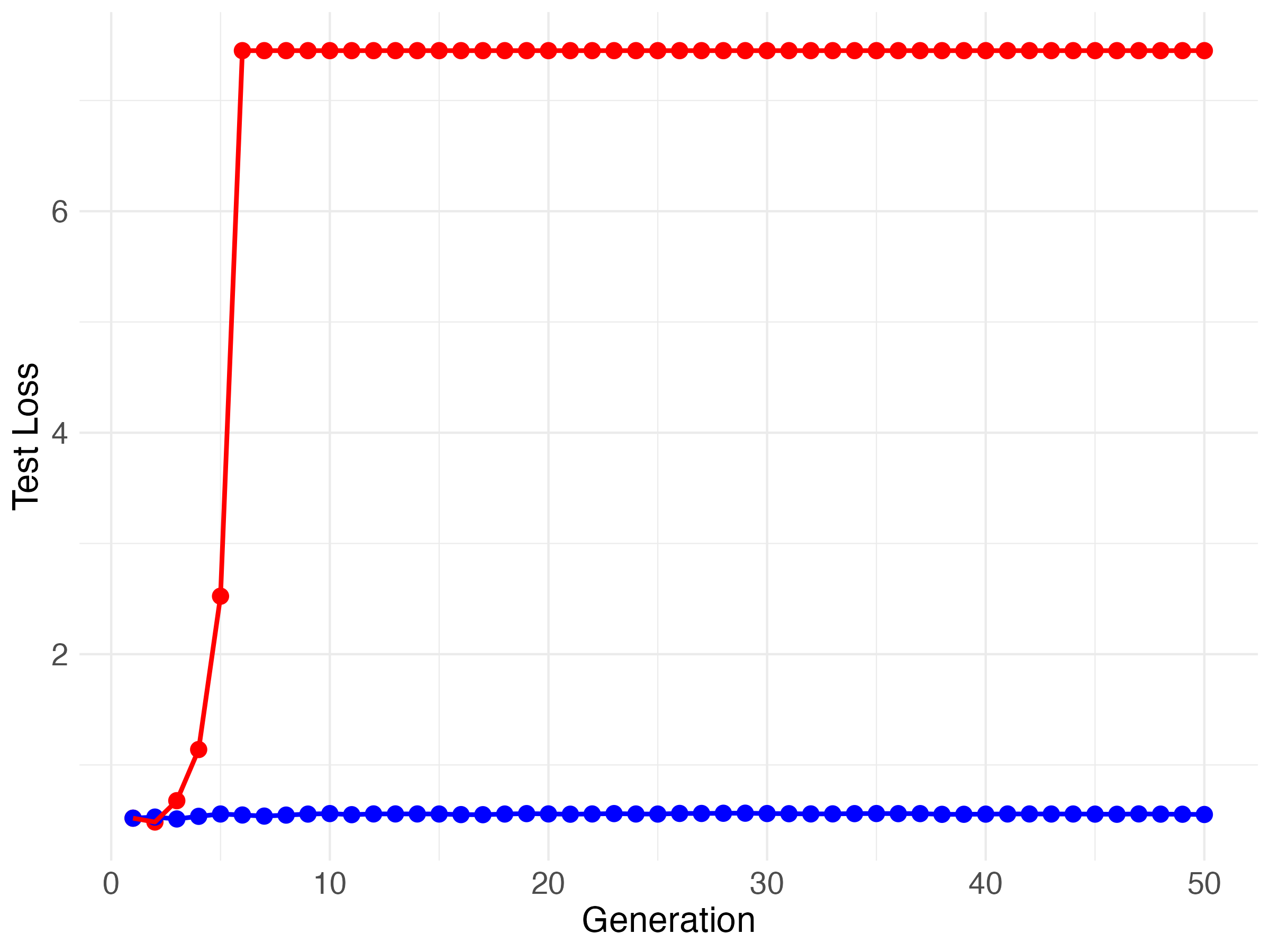}
        \subcaption{Heart}
        \label{fig:heart}
    \end{subfigure}

    \vspace{1em} % Adds some vertical space between rows

    \begin{subfigure}[b]{0.3\textwidth}
        \centering
        \includegraphics[width=\textwidth]{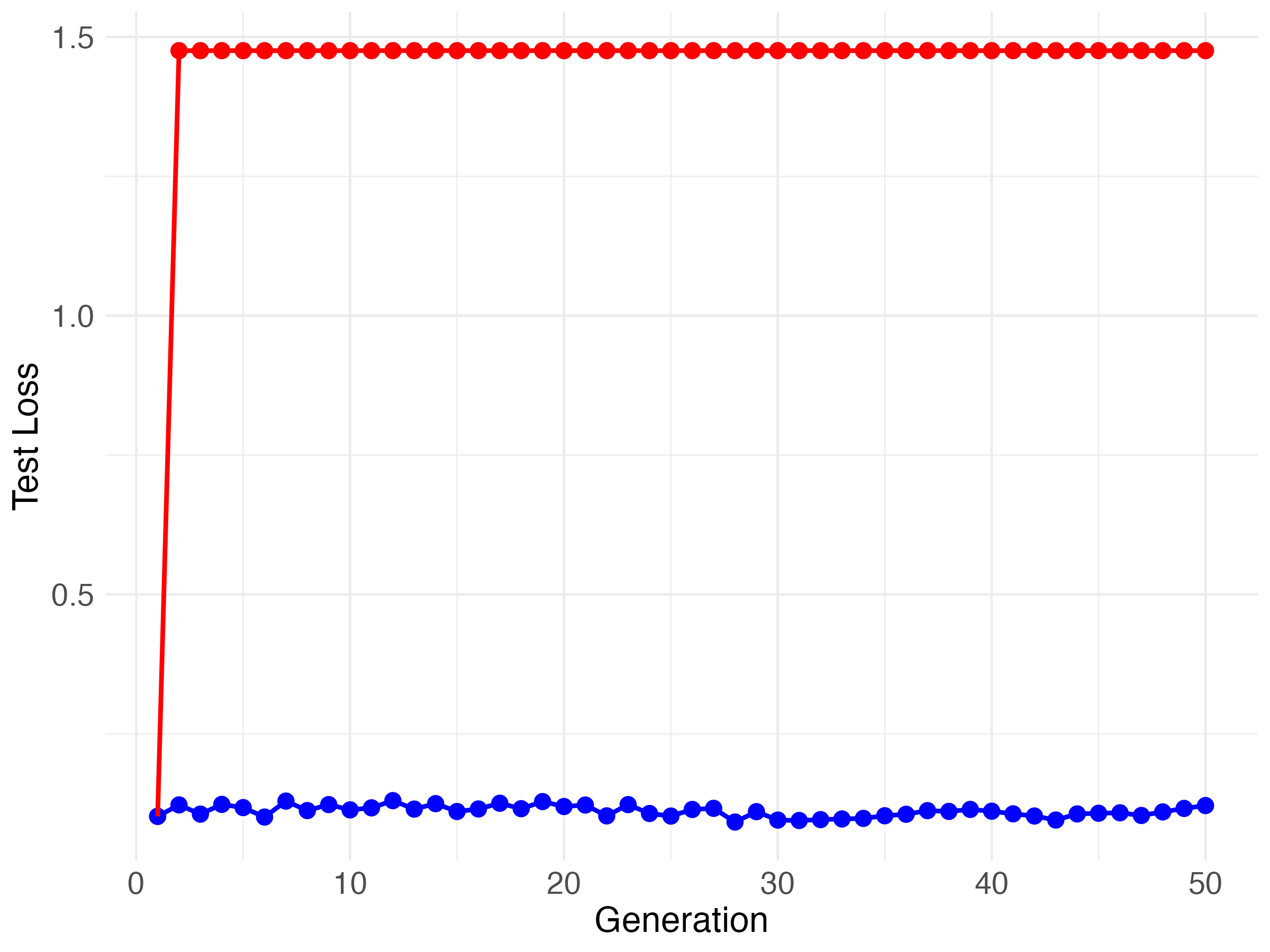}
        \subcaption{Wisconsin}
        \label{fig:wisconsin}
    \end{subfigure}
    \begin{subfigure}[b]{0.3\textwidth}
        \centering
        \includegraphics[width=\textwidth]{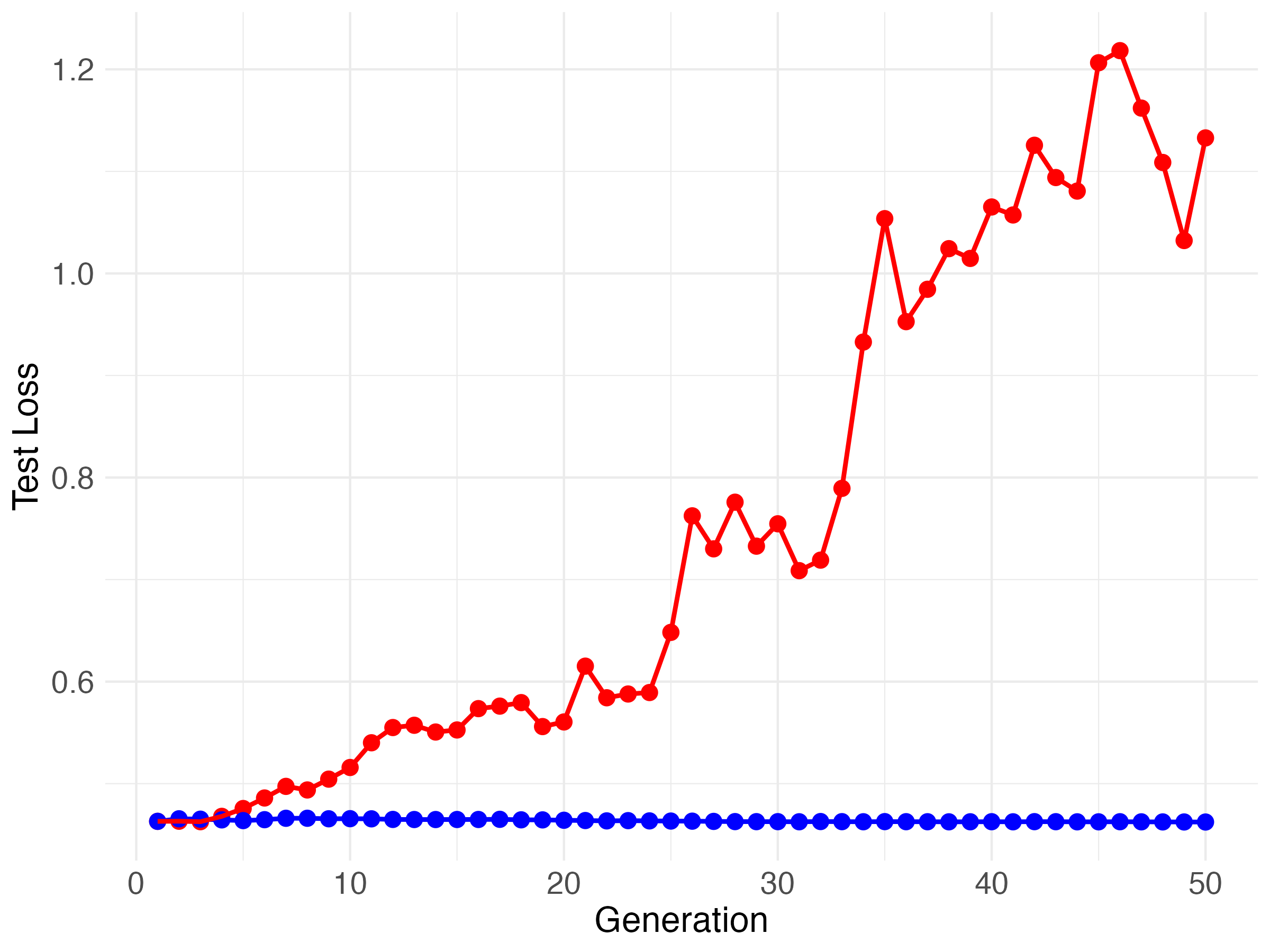}
        \subcaption{Titanic}
        \label{fig:titanic}
    \end{subfigure}

    \caption{Classification test losses on four datasets over 50 model-training iterations using iteratively fit logistic regression: (in clockwise order) Diabetes, Heart, Titanic, and Wisconsin. \color{red}Red \color{black} denotes the curve with \textit{discard} workflow, whereas \color{blue}blue \color{black} denotes the corresponding curve with the \textit{augment} workflow. We can see that the red curves are significantly higher than the blue curves in all the cases. In fact, the blue curves barely seem to increase in comparison to the red curves.}
    \label{fig:UCI}
\end{figure*}

\paragraph{Classification on Real Datasets from UCI ML Repository.} Till now, the empirical study of Model Collapse has been mostly limited to deep learning models (on language and image generation tasks primarily, hence on unstructured data). We explore the \textit{discard} vs \textit{augment} workflow to perform classification using iteratively fit logistic regression on four tabular datasets available on the UCI Machine learning Repository: Diabetes, Heart, Wisconsin Breast Cancer and Titanic. Figure \ref{fig:UCI} shows very clearly the benefit of \textit{augment} over \textit{discard}.

\begin{figure*}[h]
    \centering
    \begin{subfigure}[b]{0.3\textwidth}
        \centering
        \includegraphics[width=\textwidth]{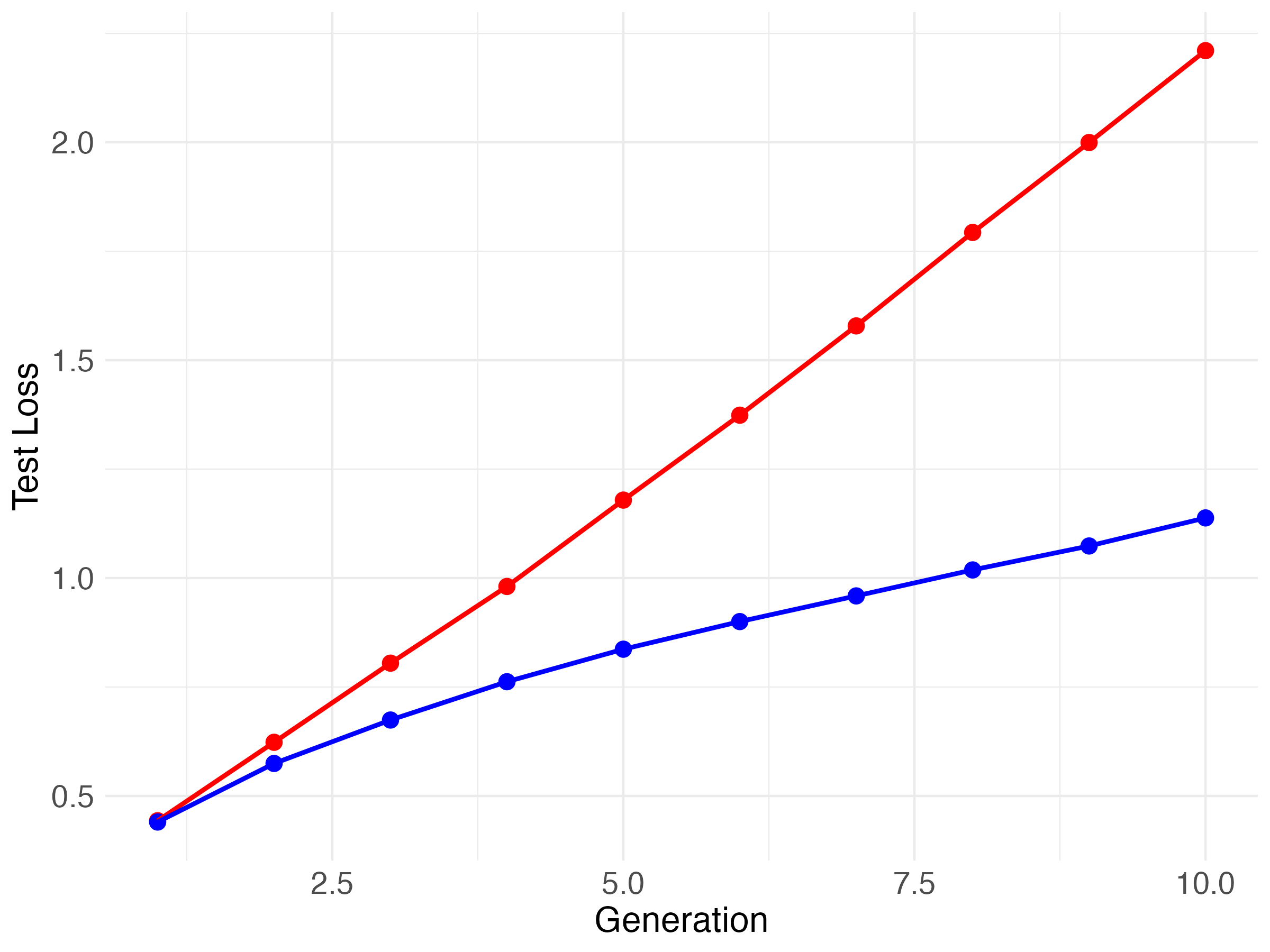}
        \subcaption{SimCLR}
        \label{fig:simclr}
    \end{subfigure}
    \begin{subfigure}[b]{0.3\textwidth}
        \centering
        \includegraphics[width=\textwidth]{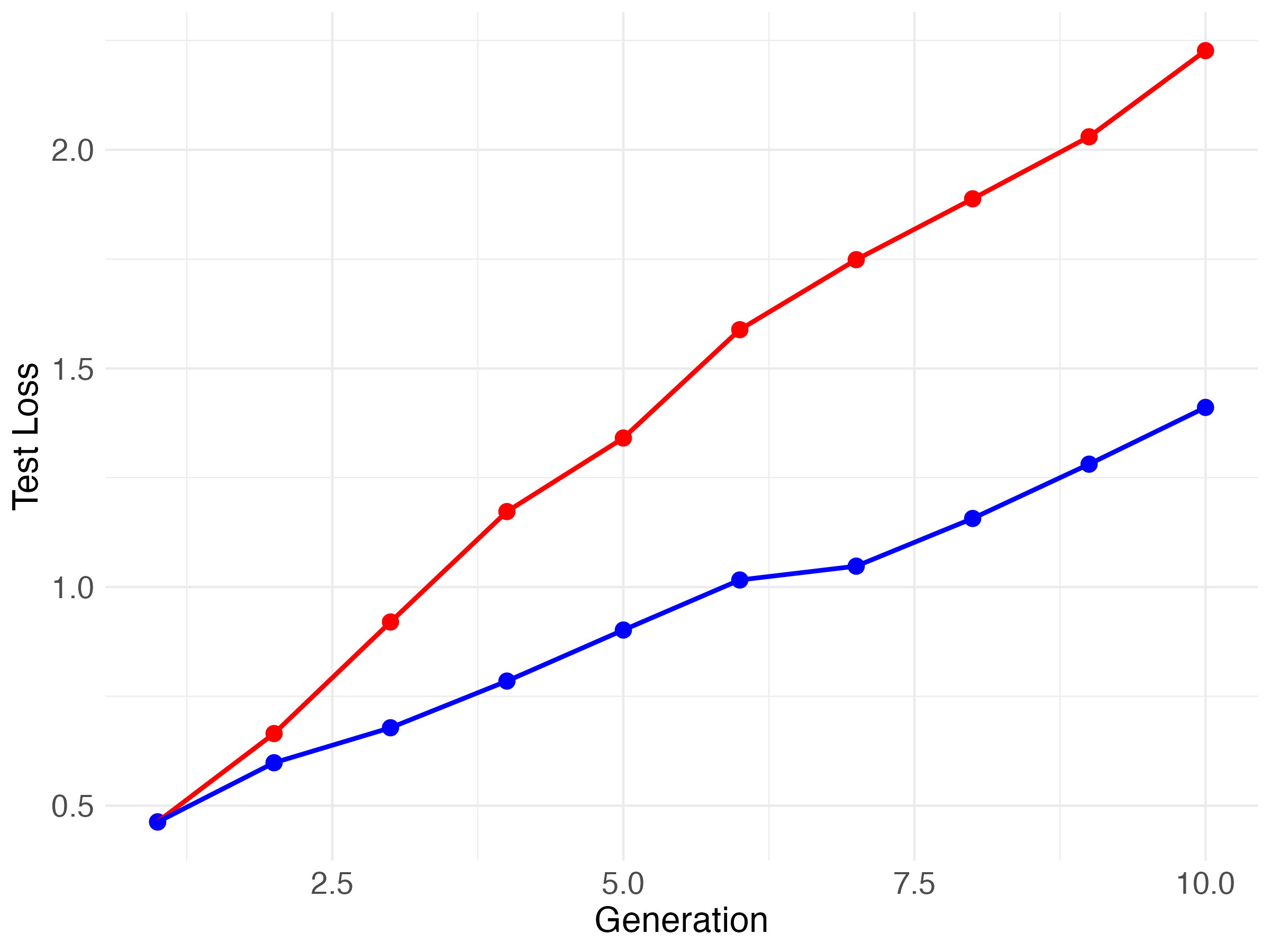}
        \subcaption{DINO}
        \label{fig:dino}
    \end{subfigure}

    \vspace{1em} % Adds some vertical space between rows

    \begin{subfigure}[b]{0.3\textwidth}
        \centering
        \includegraphics[width=\textwidth]{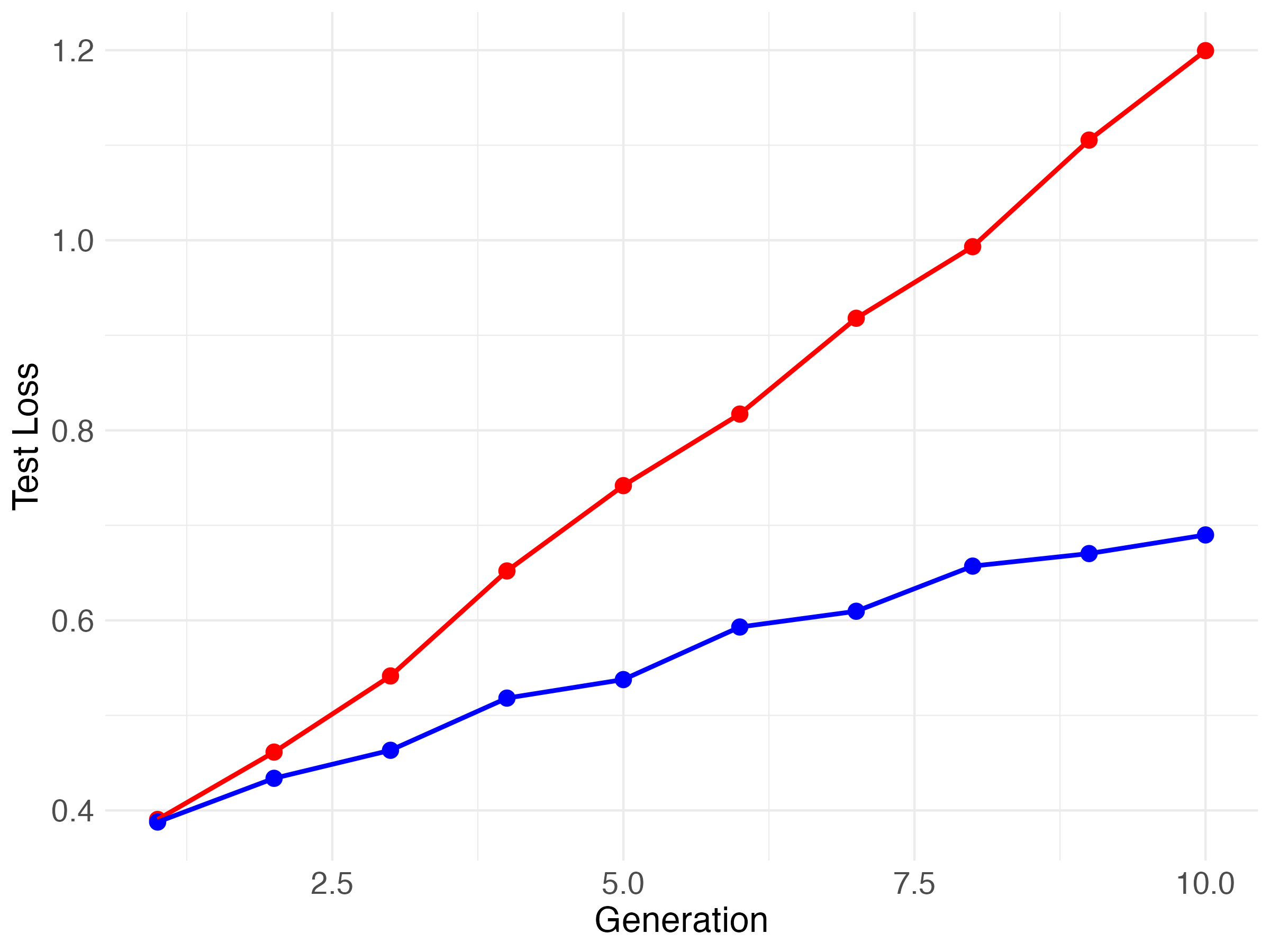}
        \subcaption{MoCoV2}
        \label{fig:mocov2}
    \end{subfigure}
    \begin{subfigure}[b]{0.3\textwidth}
        \centering
        \includegraphics[width=\textwidth]{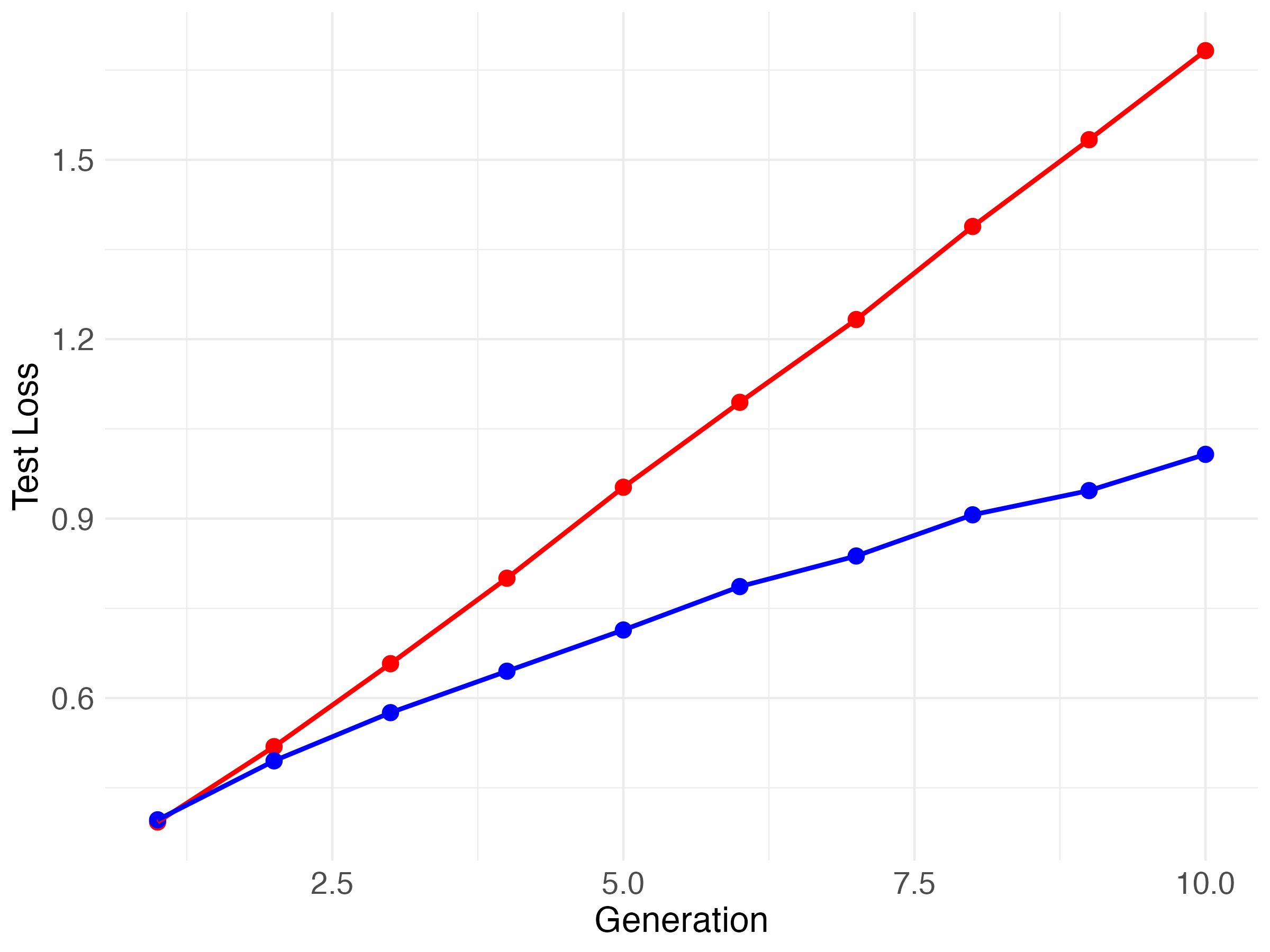}
        \subcaption{SWAV}
        \label{fig:swav}
    \end{subfigure}

    \caption{Classification test losses with four SSL trained features obtained from ResNet-50 applied on CIFAR-10. \color{red}Red \color{black}denotes \textit{discard} while \color{blue}blue \color{black} denotes \textit{augment}. The blue curves clearly increase at a slower rate.}
    \label{fig:SSL}
\end{figure*}

\paragraph{Classification with self-supervised learned features.} We obtain self-supervised learned (SSL) features on CIFAR-10 using a variety of models originally trained on ImageNet using ResNet50. The model checkpoints have been taken from \citet{lightly}, an open source GitHub repository. We train a (multinomial) logistic model using these features to classify on CIFAR-10, and experiment under both the \textit{discard} and \textit{augment} workflows. Figure \ref{fig:SSL} shows four plots from four highly impactful SSL models: SimCLR \citep{simclr}, DINO \citep{dino}, MoCoV2 \citep{moco} and SWAV \citep{swav}. In all these cases, the test loss for \textit{augment} rises more slowly across generations than does the test loss for \textit{discard}. However, we would like to highlight that the high dimensionality of the features and large sample size made it computationally infeasible to use an off-the-shelf software (e.g. \verb|LogisticRegression()|, \verb|glm()|) for fitting the logistic models. Instead, we fit the model using mini-batch gradient descent on the logistic loss, and also stop the gradient descent iterations early, and hence the estimates obtained are not necessarily what would have been obtained if we had the resources to fit an honest logistic model. Still, the benefit of \textit{augment} over \textit{discard} is already manifested.

\section{On the proof of Theorem \ref{thm:main_thm}}\label{sec:proof_structure}

\paragraph{Motivation.} On a fundamental level, proving Theorem \ref{thm:main_thm} boils down to understanding the statistical properties of the dataset obtained after generating (and augmenting) synthetic data. Clearly, our synthetic data is distributionally different from the real data; for example, the real data may consist of iid points but our synthetic data are typically not iid (as they are generated conditional upon the  a random parameter provided by a statistical estimate). Further, tracking the dependencies due to iterative model building and data generation presents significant theoretical complications. 

In the literature, so far, we have seen approaches that attempt to directly work with the shifted distribution \citep{shumailov2024ai,bertrand2024on,alemohammad2024selfconsuming,dohmatob2024model,gerstgrasser2024is,marchi2024heat,seddik2024bad}.
This lead to involved and cumbersome calculations, hard to do,  hard to interpret \-- and, seemingly, hard to enivision where they might lead even in slightly more general models.

We develop here an alternate approach. It allows us to entirely avoid working with the complexity of the intergenerational dependence of our synthetic data distributions. 
Our approach exploits fundamental ideas of statistical decision theory developed by Lucien Le Cam.
He showed that, \textit{in the locally asymptotically normal situation}, if one knows the limiting distribution of a statistic under a null hypothesis, it is possible to calculate the limiting distribution of the same under a local alternative \textit{without directly performing complicated computations under the alternative}.  Applying Le Cam's principle in 
this very different setting will be seen 
to be highly simplifying, and \textit{transparently universal}.

\paragraph{Contiguity and Le Cam's Lemma.} For two sequences of probability distributions $\{\sP_N\}_N$ and $\{\sQ_N\}_N$, we say that $\sQ_N$ \textit{is contiguous to} $\sP_N$, written as $\sQ_N\triangleleft \sP_N$, if, for any sequence of events $\{A_N\}_N$, whenever $\sP_N(A_N)\to 0$, $\sQ_N(A_N)\to0$ as well. The notion of contiguity has a rich history in statistics. It was introduced first by \citet{cam1960locally}, and it embodies the idea of \textit{transfer} of statistical properties computed under $\sP_N$ to $\sQ_N$. Perhaps the most significant application of this notion lies in the following result.

\begin{theorem}\label{thm:lecam_third_lemma}[Theorem 14.3.3 from \citet{lehmann1986testing}]
 {\bf Le Cam's Third Lemma.}
    Suppose $\sQ_N\triangleleft \sP_N$ and define $L_N(z_1,\dots,z_N)=(d\sQ_N/d\sP_N)(z_1,\cdots, z_N)$ to be the likelihood ratio. Suppose for a sequence of random variables $R_N\equiv R_N(Z_1,\cdots, Z_N)$, for every bounded continuous $g$, we know that
    \begin{align}
        \E_{\sP_N}[g(R_N,L_N)] \stackrel{N\to\infty}{\longrightarrow} \int g(r, \ell)dF(r,\ell) ,
    \end{align}for some specific distribution function $F$. Then, 
    \begin{align}
        \E_{\sQ_N}[g(R_N,L_N)] \stackrel{N\to\infty}{\longrightarrow} \int g(r, \ell) \cdot \ell \cdot  dF(r,\ell) .
    \end{align}
\end{theorem}
This fundamental result clarifies how \textit{merely} understanding the limiting behavior of the statistic under one base case $\sP_N$ (usually less challenging) is enough to completely characterize its limiting behavior under an alternate case $\sQ_N$ (usually more challenging).

\paragraph{Proof architecture for Theorem \ref{thm:main_thm}.} We will show that, for every generation $G$, the actual data distribution is contiguous to $\sP^\refer$. Under $\sP^\refer$, as discussed before, it is fairly straightforward to compute the limiting distribution. An application of Le Cam's Third Lemma (Theorem \ref{thm:lecam_third_lemma}) then yields the desired result. Details can be found in Section \ref{sec:appendix_proof_main_theorem}.

\paragraph{Intuition behind applicability of contiguity.} Contiguity and Le Cam's Third Lemma have been traditionally used in statistics to understand the asymptotics under local alternatives where the parameter deviates by a \textit{small} amount $h/\sqrt{n}$ from that under the null ($h$ is usually fixed, and $n$, usually the sample size, diverges). We are working in a seemingly different setup: the new distributions (of real-plus-synthetic data) are not explicitly shifted in this linear way. However, since each estimator $\hat\theta^\aug_G$ is $\sqrt{n}$-consistent, the new distributions \textit{do not deviate} too much from a distribution that treats all the real-plus-synthetic data as being real. Indeed, as discussed above and in the proof, contiguity holds and Le Cam's Third Lemma applies.

\section{Discussion}\label{sec:discussion}

In this work, we seek to unify the theoretical literature on model collapse by clarifying first of all the (universal) benefit of augmenting real and synthetic data, and secondly providing a useful workhorse to the data scientist who may be daunted by the numerous possible ways to fit models. A central message that we deliver is that there is no need to do model-specific calculations, and simulating a Gaussian process is all that one needs. We use this workhorse to do a comparative analysis of the \textit{discard} vs. \textit{augment} vs \textit{augment-subsample} workflows. A natural question that emerges at this point is how to optimize the weights $(\omega_{G,g,i})_{g,i}$ at generation $G$ to achieve maximal performance. Thanks to Theorem $\ref{thm:main_thm}$, this amounts to, in principle, choosing weights that minimize the variance of the asymptotic Gaussian limit, $W_\Theta(G)$. To avoid obscuring the central message, we do not pursue this line of thought in the current manuscript. Future work will study this in greater depth.

\paragraph{Code Availability.} All codes used in preparing this manuscript are available in \\
\verb|https://github.com/apd1995/model_collapse_universality|.

\section{Acknowledgement}

The authors would like to thank X.Y. Han for pointing to the GitHub repository, \citet{lightly}, containing the checkpoints of SSL models.

% \bibliography{main}

\bibliographystyle{conference}

\appendix

\section{Proof Sketch of Theorem \ref{thm:main_thm}}\label{sec:appendix_proof_main_theorem}

\subsection{Setting the stage}

In whatever follows, $\sP$ will correspond to the (actual) distribution of the $nG$ datapoints in the database $\gZ_G$ at generation $G$. This is an intractable distribution owing to complicated dependencies among the data points. Let us remind the reader that $\sP^\refer$ denotes the reference distribution on the same $nG$ data points, but which assumes all these data points are iid from the original real distribution. Following the Le Cam principle, the goal is to show contiguity of $\sP^\refer$ to $\sP$ and then LeCam's Third Lemma applies.

For every $G\geq 1$, we will find the asymptotic joint distribution of the quantity
\begin{align}
    S_{n,G} := \left(\dfrac{1}{\sqrt{n}}\sum_{i=1}^n X_{g,i}^\top(T(Y_{g,i})-\nabla A(X\theta_0)), \sqrt{n}(\hat\theta_g-\theta_0)\right)_{g=1}^G
\end{align} under $\sP^\refer$.

Using Assumption 3 that 
\begin{align}
    \sqrt{n}(\hat\theta_g-\theta_0) &= \dfrac{1}{g\sqrt{n}}\sum_{g'\leq g}\sum_{i\leq n}\psi(Z_i;\theta_0) + o_p(1)
\end{align}we conclude by the multivariate CLT (using the fact that the weights are independently chosen of the data) that $S_{n,G}\to S_G$ weakly, for a multivariate Gaussian variable $S_G$ with mean zero and covariance matrix $\Sigma_G$.

For convenience, define, for a generation $G$, the asymptotic Gaussian limits
\begin{align}
    W_T(G) &= \text{asymp.lim.}\left[\dfrac{1}{\sqrt{n}}\sum_{i=1}^n X_{G,i}^\top(T(Z_{G,i})-\nabla A(\theta_0))\right]\\
    W_\Theta(G) &= \text{asymp.lim.}\left[\sqrt{n}(\hat\theta_G-\theta_0)\right]
\end{align}

\subsection{Contiguity}

To show that $\sP$ is contiguous to $\sP^\refer$ for every generation $G\geq 1$, we will show that the likelihood ratio $L_{G,nG}$ under $\sP^\refer$ converges weakly to a random variable with mean $1$ \citep{lehmann1986testing}. 

The loglikelihood ratio is given by (setting $\hat\theta_0=\theta_0$),
\begin{align}
    &\log L_{G,nG}(Z_1,\cdots, Z_{nG}) \nonumber\\
    &= \sum_{g=1}^G\left[\sum_{i=1}^n\left((\hat\theta_{g-1}-\theta_0)^\top X_{g,i}^\top(T(Y_{g,i})-\nabla A(X_{g,i}\theta_0))-\dfrac{1}{2}(\hat\theta_{g-1}-\theta_0)^\top X_{g,i}^\top\nabla^2 A(X_{g,i}\theta_0)(\hat\theta_{g-1}-\theta_0)\right)\right]\nonumber \\
   &+ R_{G,nG}
\end{align}where
\begin{align}
    R_{G,nG} &= \sum_{g=1}^G\sum_{i=1}^n\left(A(X_{g,i}\hat\theta_{g-1})-A(X_{g,i}\theta_0)-(\hat\theta_{g-1}-\theta_0)^\top X_{g,i}^\top\nabla A(\eta(X_{g,i},\theta_0))-\right.\nonumber\\
    &\left.\dfrac{1}{2}(\hat\theta_{g-1}-\theta_0)^\top X_{g,i}^\top\nabla^2 A(X_{g,i}\theta_0)X_{g,i}(\hat\theta_{g-1}-\theta_0)\right)
\end{align}Note that under the regularity condition presented in Assumption 2, and using the fact that $H$ has finite moments of all orders (Assumption 1), $R_{G,nG}=O_p(\sum_{g=1}^G n\|\hat\theta_{g-1}-\theta_0\|^3)$. Under $\sP^\refer$, $\sqrt{n}(\hat\theta_g-\theta_0)$ is asymptotically Gaussian for each $g<G$, and therefore $R_{G,nG}=O_p(1/\sqrt{n})$.

It is clear that the loglikelihood is a continuous function of the statistic $S_{n,G}$, and hence, by the continuous mapping theorem (coupled with Slutsky's lemma that takes care of $R_{G,nG}=o_p(1)$), the limit distribution of $\log L_{G,nG}(Z_{1,1},\cdots, Z_{G,n})$ equals the distribution of
\begin{align}
    \sum_{g=1}^G\left[W_\Theta(g-1)^\top W_T(g)-\dfrac{1}{2}W_\Theta(g-1)^\top\cdot  \E(X^\top\nabla^2A(X\theta_0)X)\cdot W_\Theta(g-1)\right]
\end{align}Observe that $\Var_0(T)=\E_0(X^\top \nabla^2 A(X\theta_0)X)$. Letting $\gF_g=\sigma(Z_{1,1},\cdots, Z_{g,n})$ (the $\sigma-$field generated by the random variables up to generation $g$), we can write
\begin{align}
    \E[L_{G,nG}(Z_{1,1},\cdots, Z_{G,n})] &= \E\left[L_{G-1,n(G-1)}(Z_{1,1},\cdots, Z_{G-1,n})\times\E\left(\exp\left(W_\Theta(G-1)^\top W_T(G)\right.\right.\right.\nonumber\\
    &\left.\left.\left.-\dfrac{1}{2}W_\Theta(G-1)^\top\cdot \Var_0(T)\cdot W_\psi(G-1)\right)|\gF_{G-1}\right)\right]
\end{align}Now, recall that 
\begin{align}
    W_T(G) \sim \gN(0, \Var_0(T))
\end{align}independent of $\gF_{G-1}$. Using the moment generating function of the Gaussian variable, we get that
\begin{align}
    \E[\exp(W_\psi(G-1)^\top W_T(G)|\gF_{G-1}] &= \exp\left(\dfrac{1}{2}W_\Theta(G-1)^\top\cdot \Var_0(T)\cdot W_\Theta(G-1)\right)
\end{align}which implies that $\E[L_{G,nG}(Z_{1,1},\cdots, Z_{G,n})]=\E[L_{1,n}(Z_{1,1},\cdots, Z_{1,n})]$, which is $1$ because $L_{1,n}(Z_{1,1},\cdots, Z_{1,n})\equiv 1$ (in the first iteration, one works with real data, so there is no distribution shift). Thus, contiguity holds.

\subsection{Deriving the limiting gaussian process}

Let $p^\refer$ denote the density of the distribution $\sP^\refer$. Note that we can write the joint density of the variables under $\sP^\refer$ till generation $G$ as
\begin{align*}
    p^\refer(W_T^\refer(g),W_\Theta^\refer(g),g\leq G) &= \left(\prod_{g=1}^G p^\refer(W_T^\refer(g))\right)\times \\
    &\prod_{G'=1}^Gp^\refer(W_\Theta^\refer(G')|W_T^\refer(G'),\{W_\Theta^\refer(g),W_T^\refer(g)\}_{g<G'})
\end{align*}This is because $W_T^\refer(1),\cdots, W_T^\refer(G)$ are all independent $\gN(0,\gV_T)$.

% We now analyze the actual distribution $\sP^{actual}$. We will find its density $p^{actual}$. 
Recall the convergence of the likelihood ratio
\begin{align*}
    L_{G,nG}(Z_1,\cdots, Z_{nG}) &\to \exp\left(\sum_{g=1}^G\left[W_\Theta(g-1)^\top W_T(g)-\dfrac{1}{2}W_\Theta(g-1)^\top\cdot  \gV_T\cdot W_\Theta(g-1)\right]\right)\\
    &=: L_{G,\infty}(W_\Theta(g),W_T(g),g\leq G)
\end{align*}
According to LeCam's Third Lemma, the joint density of the limiting variables $(W_T(g),W_\Theta(g))_{g\leq G}$ equals
\begin{align*}
    p(W_\Theta(g),W_T(g),g\leq G) &= p^\refer(W_\Theta(g),W_T(g),g\leq G)\times L_{G,\infty}(W_\Theta(g),W_T(g),g\leq G)\\
    &=\prod_{G'=1}^Gp^\refer(W_\Theta^\refer(G')|W_T^\refer(G'),\{W_\Theta^\refer(g),W_T^\refer(g)\}_{g<G'})\\
    &\times \left(\prod_{g=1}^G p^\refer(W_T(g))\right)\times L_{G,\infty}(W_\Theta(g),W_T(g),g\leq G)
\end{align*}Plugging in $p^\refer(W_T(g))=\exp(-W_T(g)^\top \gV_T^{-1} W_T(g)/2)/\sqrt{\det(2\pi\gV_T)}$, we notice that
\begin{align*}
    &\left(\prod_{g=1}^G p^\refer(W_T(g))\right)\times L_{G,\infty}(W_\Theta(g),W_T(g),g\leq G) \\
    &= \prod_{g=1}^G \dfrac{1}{\sqrt{\det(2\pi\gV_T)}}\exp\left(-\dfrac{1}{2}(W_T(g)-\gV_T W_\Theta(g-1))^\top \gV_T^{-1}(W_T(g)-\gV_T W_\Theta(g-1))\right)\\
    &=: \prod_{g=1}^G p(W_T(g)|W_\Theta(g-1))
\end{align*}As a result, we get that the limiting joint density equals
\begin{align*}
        &p(W_\Theta(g),W_T(g),g\leq G) \\
        &= \prod_{G'=1}^G\left(p(W_T(G')|W_\Theta(G'-1))\times p^\refer(W_\Theta(G')|W_T(G'),\{W_\Theta(g),W_T(g)\}_{g<G'}\right)
\end{align*}
This tells us that sequentially, at each generation $G$, $W_T(G)$ evolves as a Gaussian variable with mean $\gV_TW_\Theta(G-1)$ and variance $\gV_T$, and $\gW_\Theta(G)$ evolves just as in the reference distribution conditional on $\{W_T(G),(W_\Theta(g),W_T(g))_{g<G}\}$. This proves Theorem \ref{thm:main_thm}.

\section{Implications on workflows}\label{sec:implications}

Before delving into specific cases, let us bring together some common notations. Denote for every $G\geq 1$, $\gV_T:=\Var(W_T^\refer(G))=\E_0[X^\top \nabla^2A(X\theta_0)X]$ and
\begin{align}
    \gV_\Theta &= (\E_0[\nabla^2 L(Z;\theta_0)])^{-1}\E_0[\nabla L(Z;\theta_0)(\nabla L(Z;\theta_0))^\top](\E_0[\nabla^2 L(Z;\theta_0)])^{-1}
\end{align}Further, from standard exponential family properties,
\begin{align}
    \Cov_0(\nabla L(Z;\theta_0),X^\top T(Y)) &= \E_0[\nabla^2 L(Z;\theta_0)]
\end{align}

\subsection{Proof of Lemma \ref{lemma:discard}}

In this case, we get $\Var(W_\Theta^\refer(G))=\gV_\Theta$ for every $G$. For any $G'\neq G$, all covariances are $0$. Putting these together,
\begin{align}
    \Cov(W_T^\refer(G),W_T^\refer(G')) &= \gV_T\sI_{G=G'},\\
    \Cov(W_\Theta^\refer(G),W_\Theta^\refer(G')) &= \gV_\Theta\sI_{G=G'},\\
    \Cov(W_\Theta^\refer(G),W_T^\refer(G')) &= I\sI_{G= G'}
\end{align}Under $\sP^\refer$, given
\begin{align*}
    \{W_T^\refer(G)=x_T(G),W_\Theta^\refer(G-1)=x_\Theta(G-1),\cdots, W_T^\refer(1)=x_T(1)\}
\end{align*} the conditional mean of $W_\Theta^\refer(G)$ equals 
\begin{align}
    M_\Theta(G) &= \gV_T^{-1}x_T(G)
\end{align}and the conditional variance equals
\begin{align}
    V_\Theta(G) &= \gV_\Theta - \gV_T^{-1}
\end{align}As a result, using Theorem \ref{thm:main_thm},
\begin{align}
    \Var(W_\Theta(G)) &= \gV_\Theta-\gV_T^{-1}+ \gV_T^{-1}\Var(W_T(G))\gV_T^{-1} \\
    &= \gV_\Theta-\gV_T^{-1}+ \gV_T^{-1}\cdot (\gV_T+\gV_T\Var(W_\Theta(G-1))\gV_T)\cdot\gV_T^{-1}\\
    &= \gV_\Theta + \Var(W_\Theta(G-1))
\end{align}and thus we get $\Var(W_\Theta(G)) = G \times \gV_\Theta$.

\subsection{Proof of Lemma \ref{lemma:augment}}

In this case, we get $\Var(W_\Theta^\refer(G))=\gV_\Theta/G$. Further, we calculate that for all $1\leq G'\leq G$,
\begin{align}
    \Cov(W_T^\refer(G),W_T^\refer(G')) &= \gV_T\sI_{G=G'},\\
    \Cov(W_\Theta^\refer(G),W_\Theta^\refer(G')) &= \dfrac{\gV_\Theta}{G},\\
    \Cov(W_\Theta^\refer(G),W_T^\refer(G')) &= \dfrac{I}{G}\sI_{G\geq G'}
\end{align}

Now we apply Theorem \ref{thm:main_thm}. Under $\sP^\refer$, given
\begin{align*}
    \{W_T^\refer(G)=x_T(G),W_\Theta^\refer(G-1)=x_\Theta(G-1),\cdots, W_T^\refer(1)=x_T(1)\}
\end{align*} the conditional mean of $W_\Theta^\refer(G)$ can be computed to be
\begin{align}
    M_\Theta(G) &= \dfrac{1}{G}\gV_T^{-1}x_T(G) + \left(\dfrac{G-1}{G}\right)x_\Theta(G-1)
\end{align}and the conditional variance is
\begin{align}
    V_\Theta(G) &= \dfrac{\gV_\Theta-\gV_T^{-1}}{G^2}
\end{align}
As a result, we obtain,
\begin{align}
    \Var(W_\Theta(G)) &= \Var(M_\Theta(G)) + V_\Theta(G) \nonumber\\
    &= \Var\left(\dfrac{1}{G}\gV_T^{-1}W_T(G)+ \left(\dfrac{G-1}{G}\right)W_\Theta(G-1)\right) + V_\Theta(G)
\end{align}Writing $W_T(G)=W_T^\refer(G)+\gV_TW_\Theta(G-1)$ where $W_T^\refer\sim \gN(0, \gV_T)$ independent of everything in the past (and hence in particular $W_\Theta(G-1)$), we get
\begin{align}
    \Var(W_\Theta(G)) &= \Var\left(\dfrac{1}{G}\gV_T^{-1}W_T^\refer(G) + W_\Theta(G-1)\right) \nonumber \\
    &= \dfrac{\gV_T^{-1}}{G^2} + \Var(W_\Theta(G-1)) + \dfrac{\gV_\Theta -\gV_T^{-1}}{G^2}\\
    &= \dfrac{\gV_\Theta}{G^2} + \Var(W_\Theta(G-1))
\end{align}Unfolding this recursion and realizing that $Var(W_\Theta(1))=\gV_\Theta$, we get that
\begin{align}
    \Var(W_\Theta(G)) &= \gV_\Theta\times\left(\sum_{g=1}^G\dfrac{1}{g^2}\right)
\end{align}which concludes the proof.

\subsection{On the \textit{augment-subsample} workflow}

The \textit{augment-subsample} workflow considers (random) weights $(\omega_{G,g,i})_{g,i}$ that are a uniformly random draw from the set of $nG-$dimensional binary $(0/1)$ tuples each having exactly $n$ 1's. In this workflow, we calculate that for all $1\leq G'\leq G$,
\begin{align}
    \Cov(W_T^\refer(G),W_T^\refer(G')) &= \gV_T\sI_{G=G'},\\
    \Cov(W_\Theta^\refer(G),W_\Theta^\refer(G')) &= \gV_\Theta,\\
    \Cov(W_\Theta^\refer(G),W_T^\refer(G')) &= \dfrac{I}{G}\sI_{G\geq G'}
\end{align}As mentioned previously, it seems to be difficult to obtain exact analytic expression for $\Var(W_\Theta(G))$. However, the specifications of these variances and covariances make it simple to simulate the Gaussian process (Figure \ref{fig:gaussian_limits} has been obtained in this way).

\subsection{Proofs of Lemmas \ref{lemma:discard_test} and \ref{lemma:augment_test}}\label{sec:appdx_predictive_loss}

For a point $Z=(X,Y)$ generated from $\sP_0$, the log-likelihood ratio equals
\begin{align}
    \log(p(Y|X\hat\theta)/p(Y|X\theta_0)) &= (\hat\theta-\theta_0)^\top X^\top T(Y) - (A(X\hat\theta)-A(X\theta_0))
\end{align}which, by Taylor expansion, we will write as
\begin{align}
    \log(p(Y|X\hat\theta)/p(Y|X\theta_0)) &= (\hat\theta-\theta_0)^\top X^\top (T(Y) - \nabla A(X\theta_0)) - \dfrac{1}{2} (\hat\theta-\theta_0)^\top X^\top \nabla^2 A(X\theta_0) X(\hat\theta - \theta_0) \\
    & + \text{Rem}
\end{align}where $|\text{Rem}|=o_p(\|\hat\theta-\theta_0\|^2)$.  By Lemmas \ref{lemma:discard} and \ref{lemma:augment}, $\sqrt{n}(\hat\theta_G^{dis}-\theta_0)$ and $\sqrt{n}(\hat\theta_G^{aug}-\theta_0)$ are tight, and thus for both cases, $\hat\theta_G\stackrel{p}{\to}\theta_0$. By the regularity condition in Assumption 2, and by the fact that $H$ has finite moments of all orders (Assumption 1), we get $|\text{Rem}|=O_p(\|\hat\theta-\theta_0\|^3)=O_p(n^{-3/2})$.

Thus we get, formally, under regularity conditions,
\begin{align}
    \E D_{KL}(\hat\theta\|\theta_0) &\approx \dfrac{1}{2}\Tr(\Var_0(T)\E(\hat\theta-\theta_0)(\hat\theta-\theta_0)^\top) + o(1/n)
\end{align}

Identifying $\sqrt{n}(\hat\theta_G^{dis}-\theta_0)$ and $\sqrt{n}(\hat\theta_G^{aug}-\theta_0)$ with their (mean zero) Gaussian limits $W_\Theta^{dis}(G)$ and $W_\Theta^{aug}(G)$ (whose variances have been derived in Lemmas \ref{lemma:discard} and \ref{lemma:augment}, we get
\begin{align*}
    n\E((\hat\theta_G^{dis}-\theta_0)(\hat\theta_G^{dis}-\theta_0)^\top) \approx G\times \Var(W_\Theta(1))
\end{align*}and
\begin{align*}
    n\E((\hat\theta_G^{aug}-\theta_0)(\hat\theta_G^{aug}-\theta_0)^\top) \approx \left(\sum_{g=1}^G \dfrac{1}{g^2}\right)\times\Var(W_\Theta(1))
\end{align*}Realizing that
\begin{align}
    n\E((\hat\theta_1-\theta_0)(\hat\theta_1-\theta_0)^\top) \approx \Var(W_\Theta(1))
\end{align}we get the desired answer.

\section{Further Experimental Results}\label{sec:appdx_exp_details}

\subsection{Test accuracies for UCI ML datasets}

\begin{figure*}[h]
    \centering
    \begin{subfigure}[b]{0.3\textwidth}
        \centering
        \includegraphics[width=\textwidth]{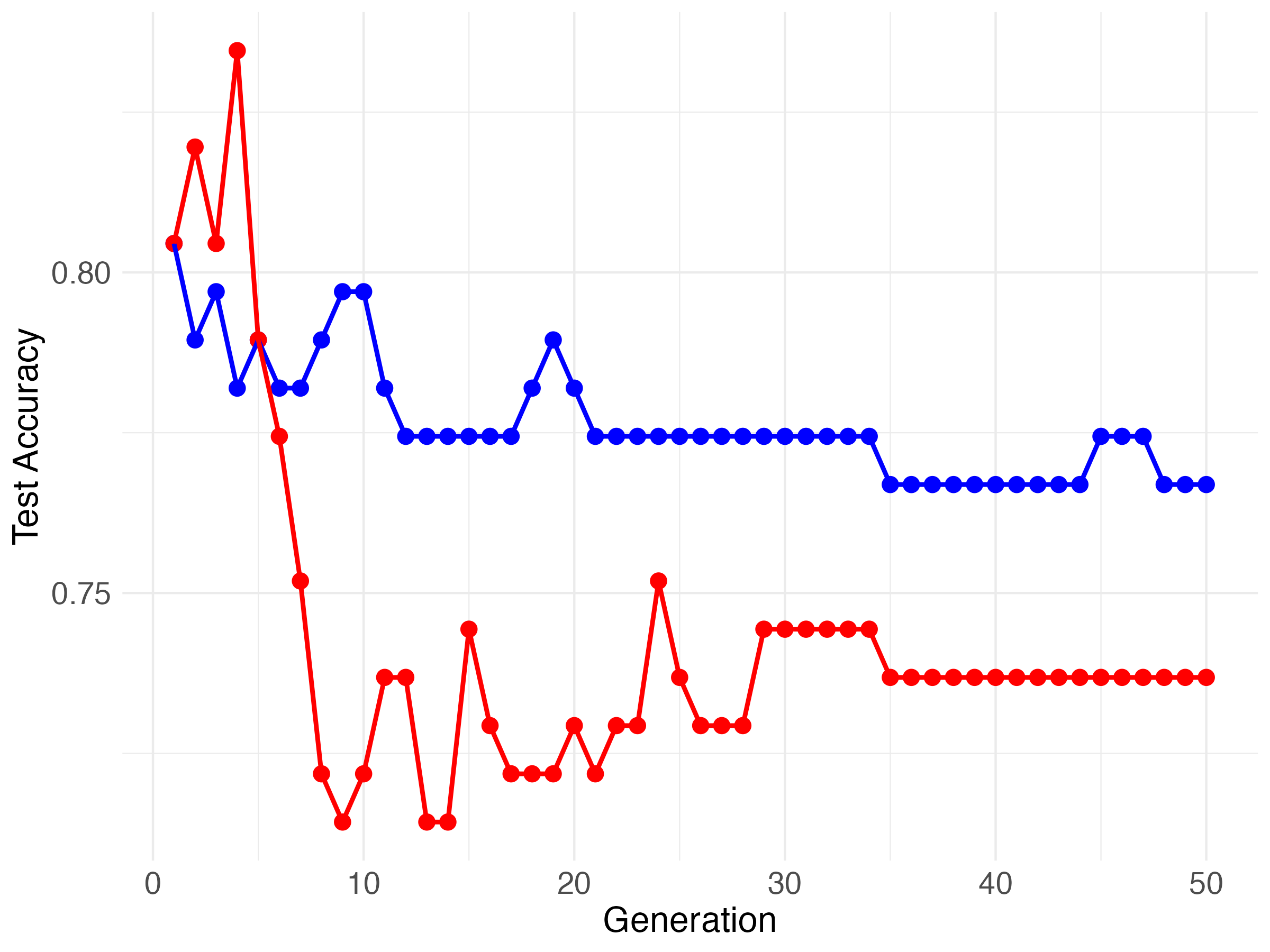}
        \subcaption{Diabetes}
    \end{subfigure}
    \begin{subfigure}[b]{0.3\textwidth}
        \centering
        \includegraphics[width=\textwidth]{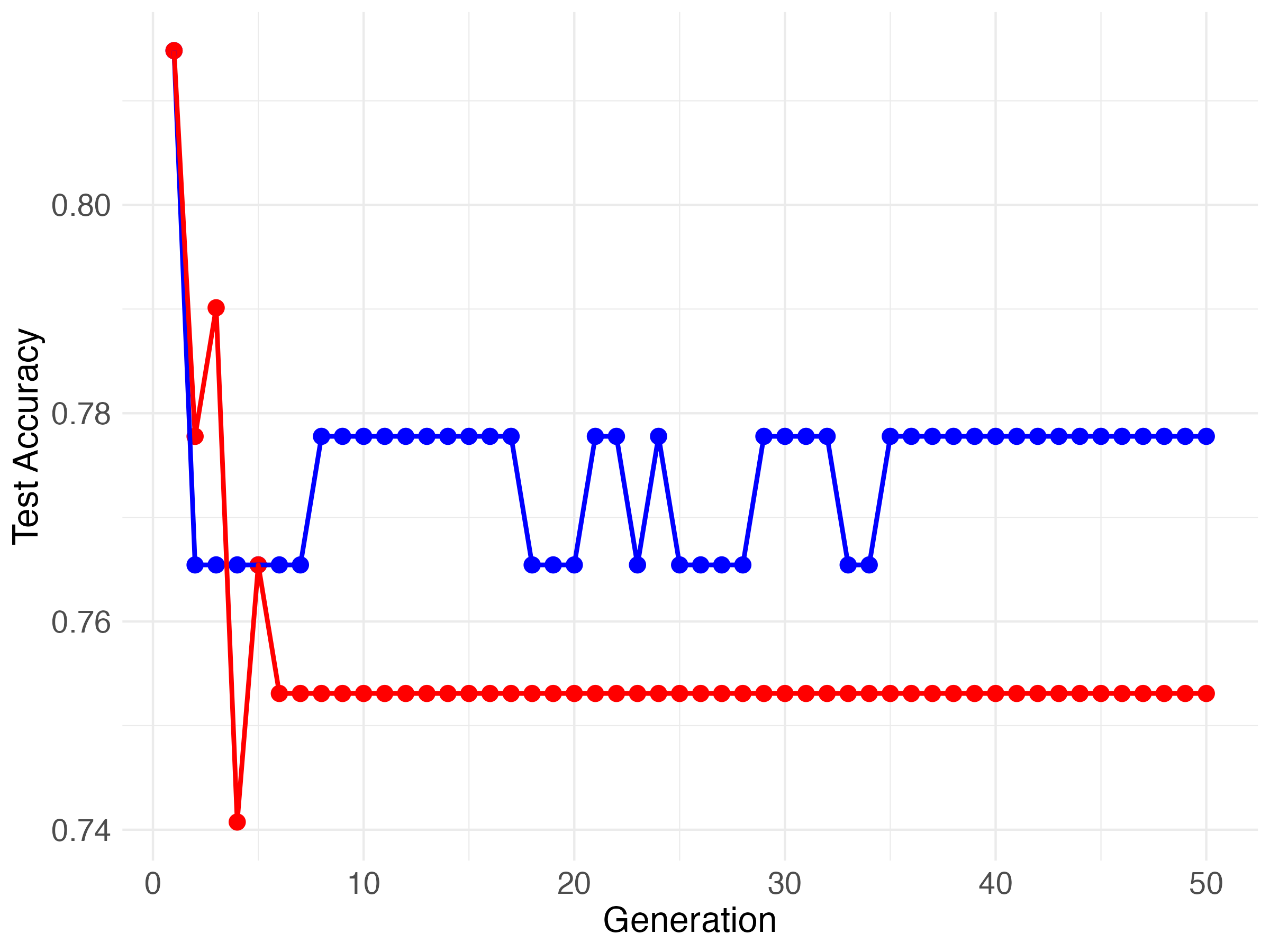}
        \subcaption{Heart}
    \end{subfigure}

    \vspace{1em} % Adds some vertical space between rows

    \begin{subfigure}[b]{0.3\textwidth}
        \centering
        \includegraphics[width=\textwidth]{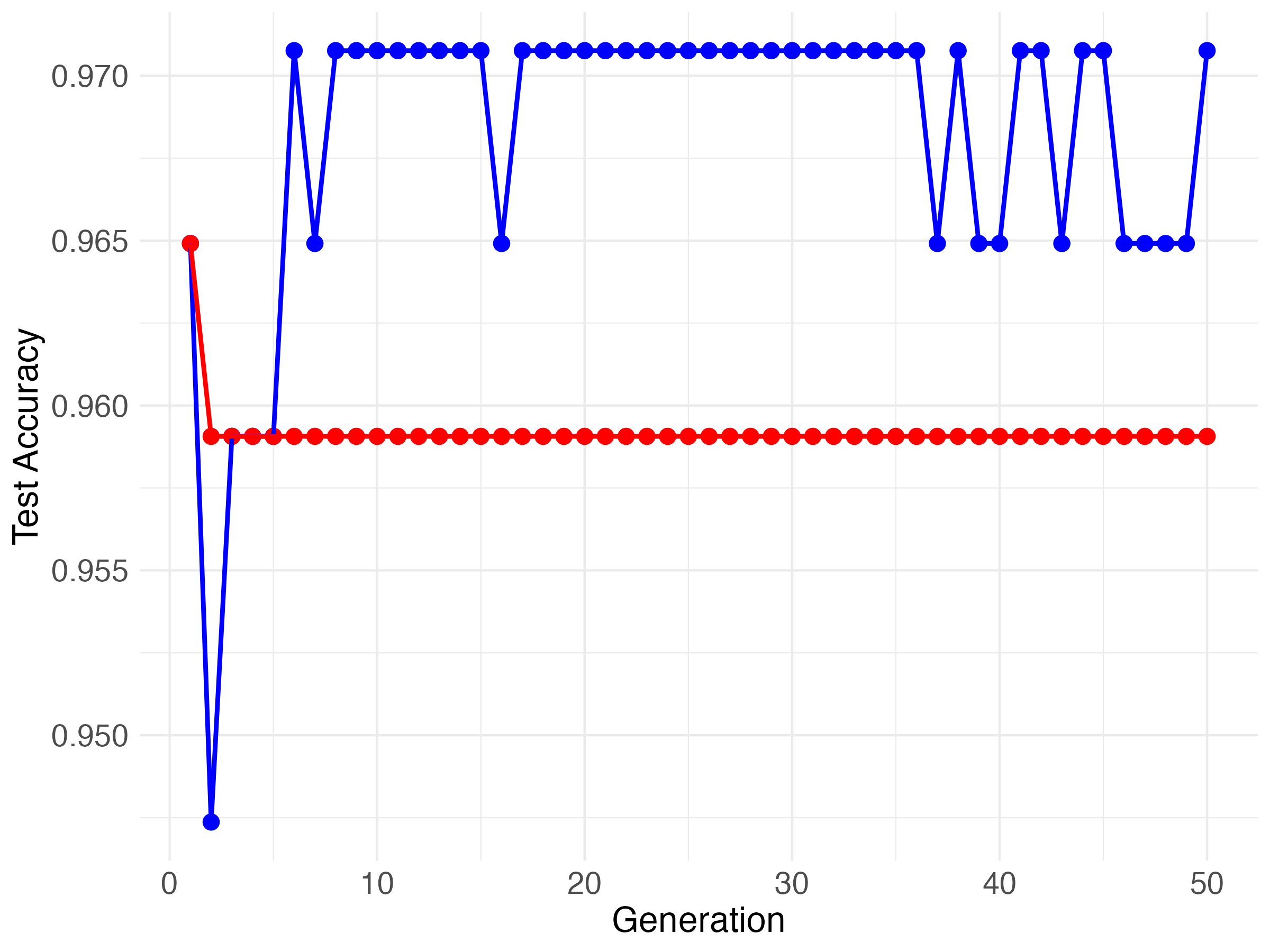}
        \subcaption{Wisconsin}
    \end{subfigure}
    \begin{subfigure}[b]{0.3\textwidth}
        \centering
        \includegraphics[width=\textwidth]{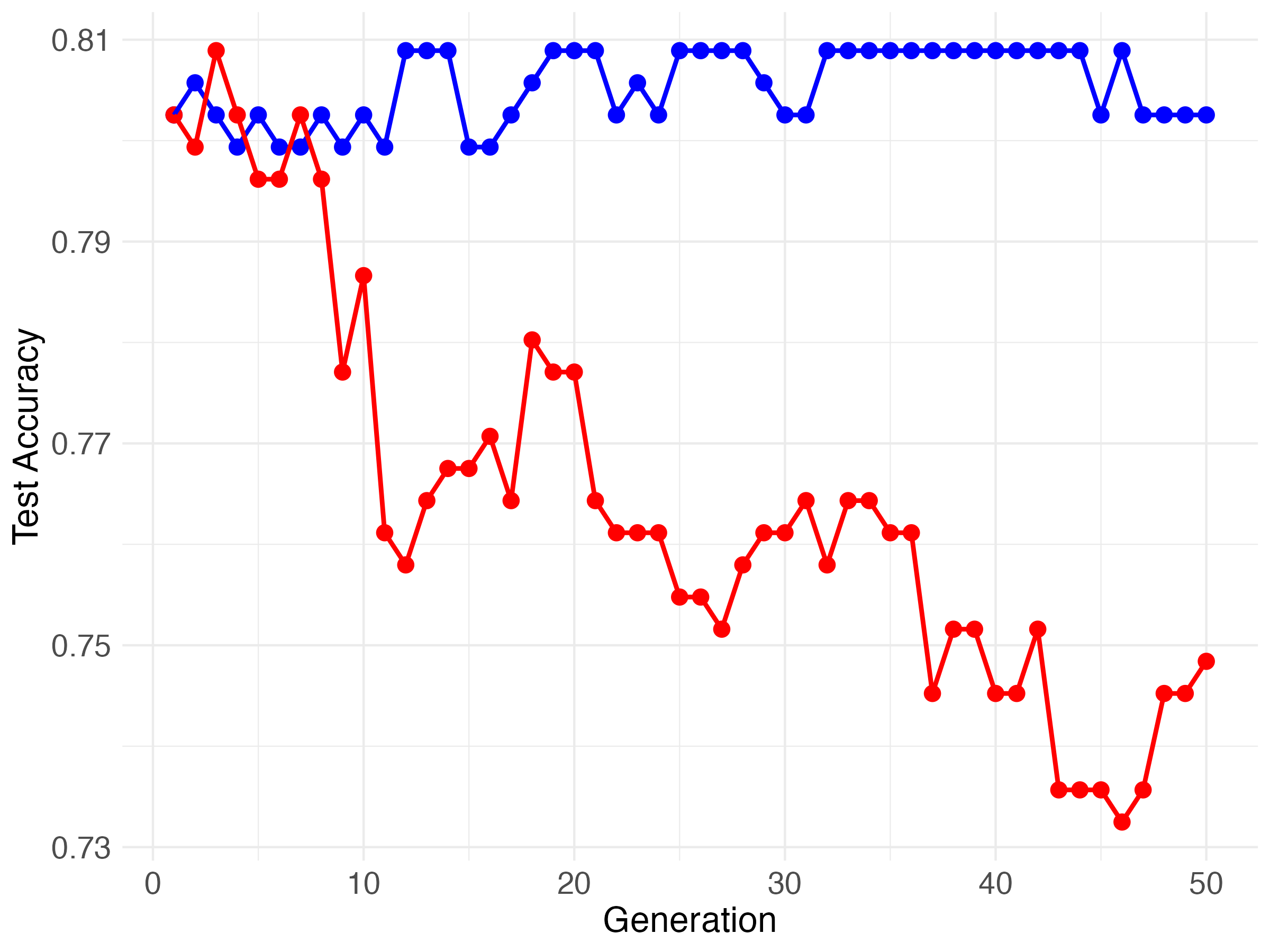}
        \subcaption{Titanic}
    \end{subfigure}

    \caption{Classification test accuracies on four datasets over 50 model-training iterations using iteratively fit logistic regression: (in clockwise order) Diabetes, Heart, Titanic, and Wisconsin. \color{red}Red \color{black} denotes the curve with \textit{discard} workflow, whereas \color{blue}blue \color{black} denotes the corresponding curve with the \textit{augment} workflow. We can see that the red curves detoriate much faster than the blue curves.}
    \label{fig:UCI_accuracy}
\end{figure*}

In Section \ref{sec:experiments}, we have presented the behavior of the test losses obtained from iterative logistic fit on four UCI ML datasets. We now provide the test accuracy plots in Figure \ref{fig:UCI_accuracy}. We find that in all the cases, the \textit{discard} accuracies decay much faster than the corresponding \textit{augment} accuracies. In the initial model fitting generations, sometimes, interestingly, \textit{discard} outperforms \textit{augment}. This is an observation that inspires further exploration, and more experimental results will be provided in subsequent versions towards understanding this.

\subsection{Test losses for SSL trained iterative logistic fit}

\begin{figure*}[h]
    \centering
    \begin{subfigure}[b]{0.3\textwidth}
        \centering
        \includegraphics[width=\textwidth]{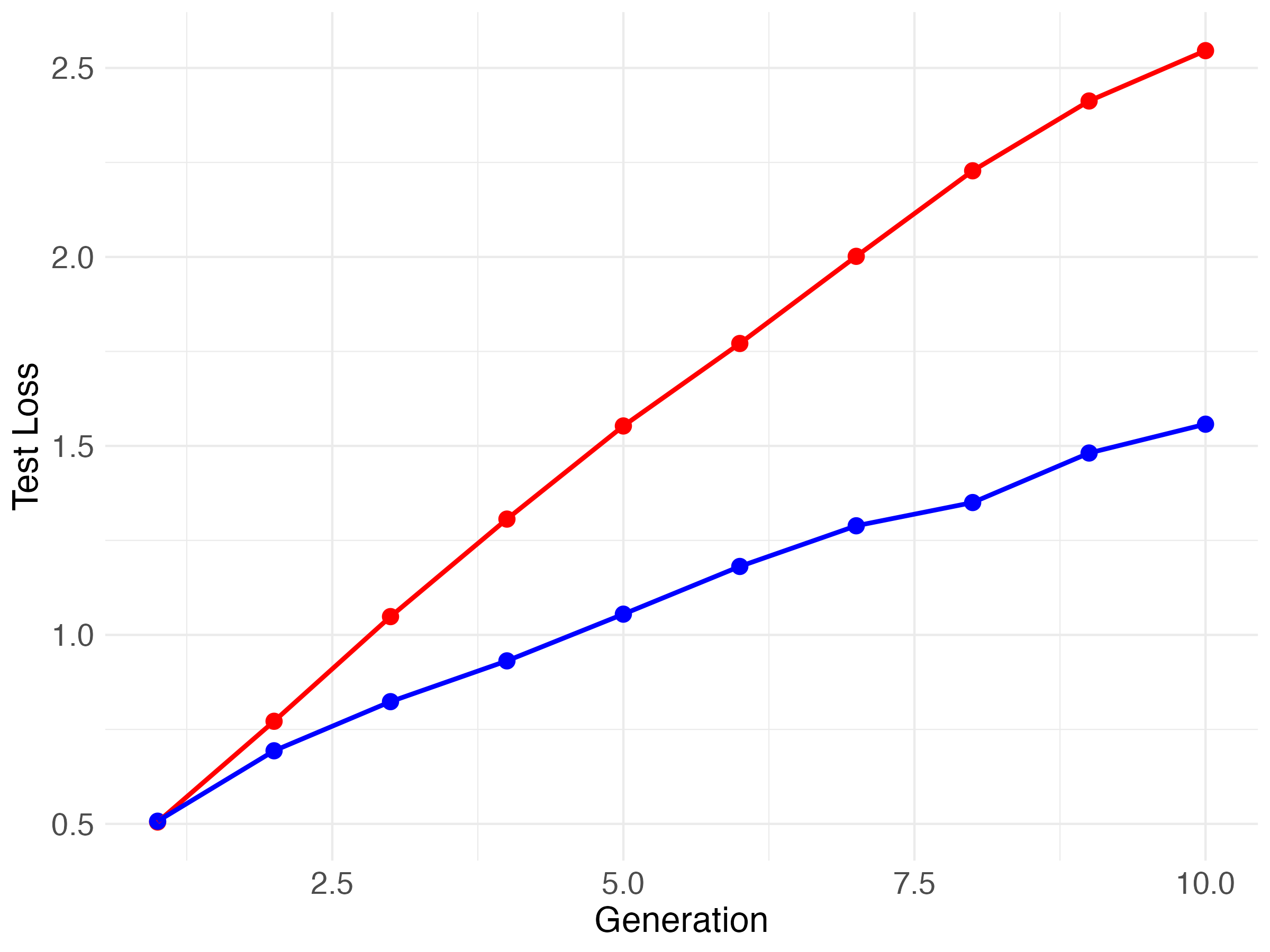}
        \subcaption{Barlow Twins}
    \end{subfigure}
    \begin{subfigure}[b]{0.3\textwidth}
        \centering
        \includegraphics[width=\textwidth]{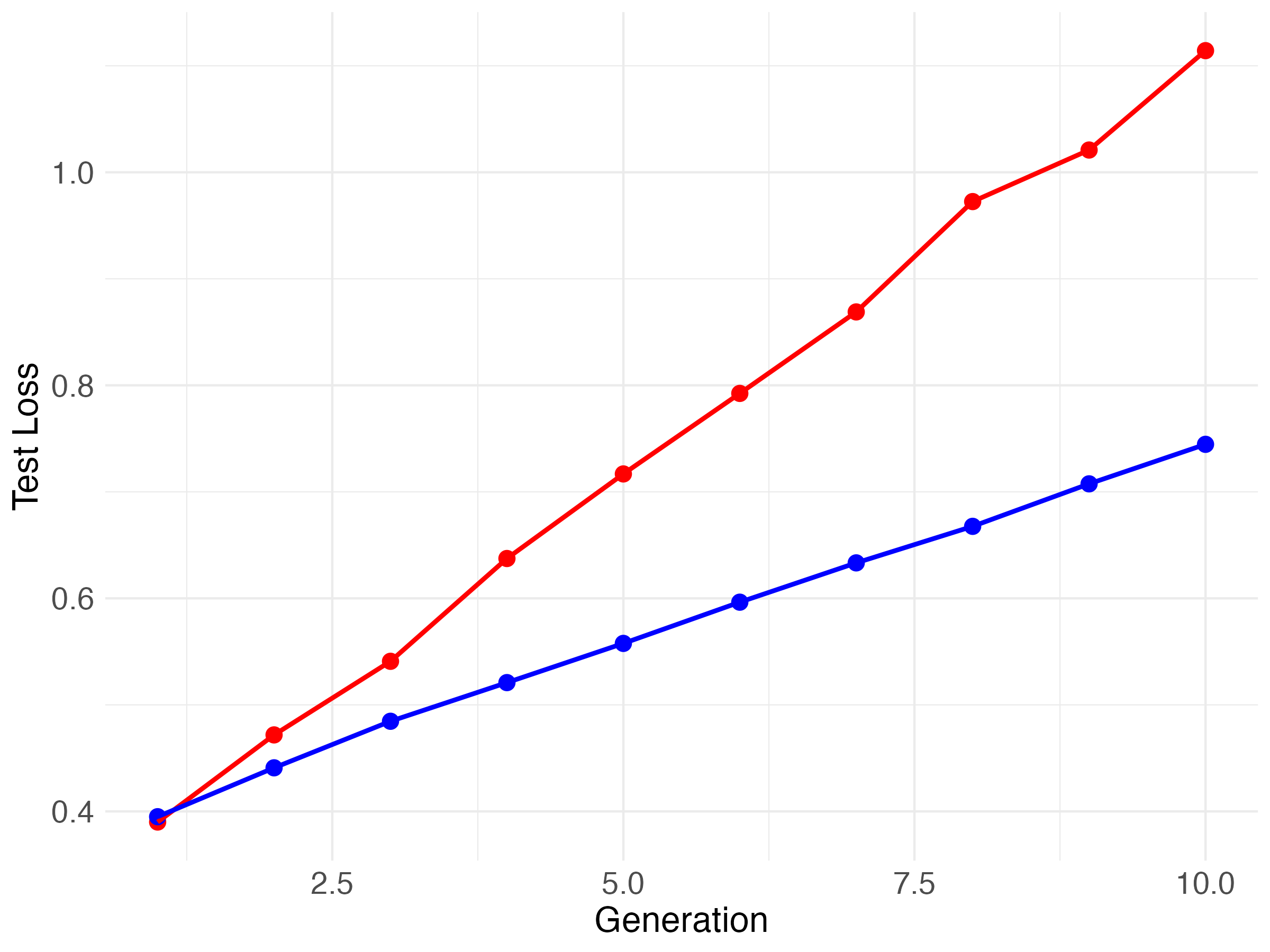}
        \subcaption{Byol}
    \end{subfigure}

    \vspace{1em} % Adds some vertical space between rows

    \begin{subfigure}[b]{0.3\textwidth}
        \centering
        \includegraphics[width=\textwidth]{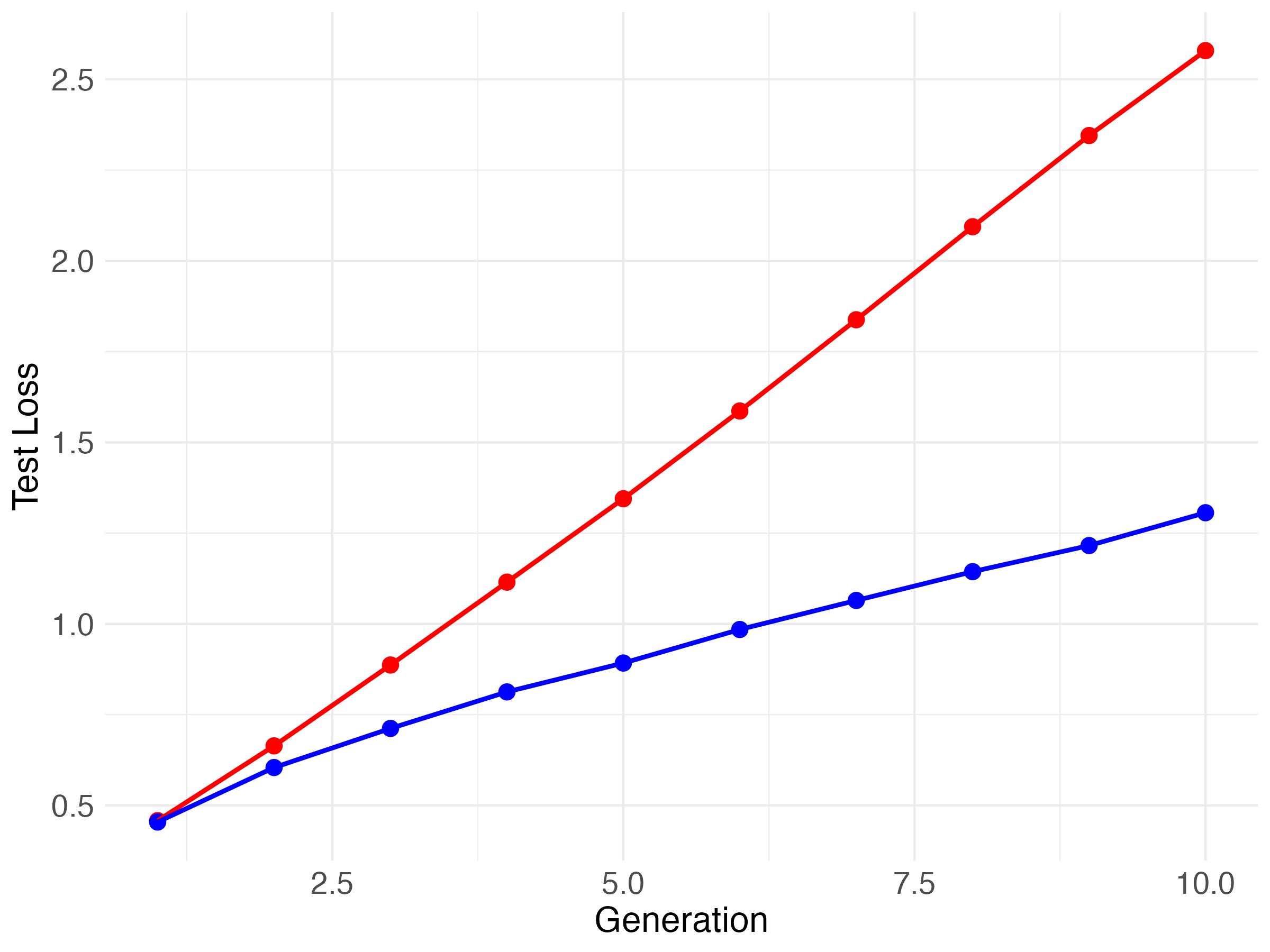}
        \subcaption{DCL}
    \end{subfigure}
    \begin{subfigure}[b]{0.3\textwidth}
        \centering
        \includegraphics[width=\textwidth]{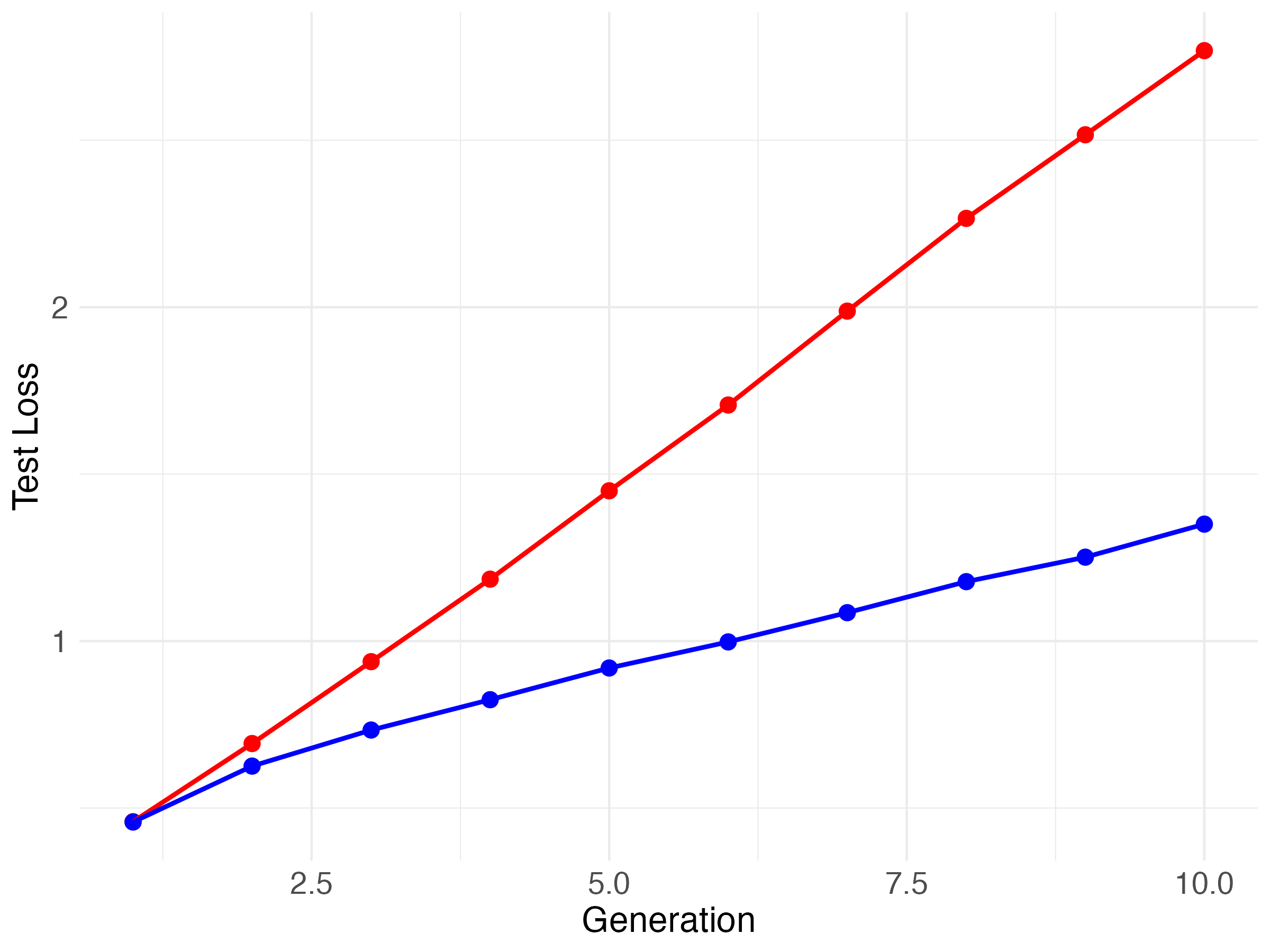}
        \subcaption{DCLW}
    \end{subfigure}

    \vspace{1em} % Adds some vertical space between rows

    \begin{subfigure}[b]{0.3\textwidth}
        \centering
        \includegraphics[width=\textwidth]{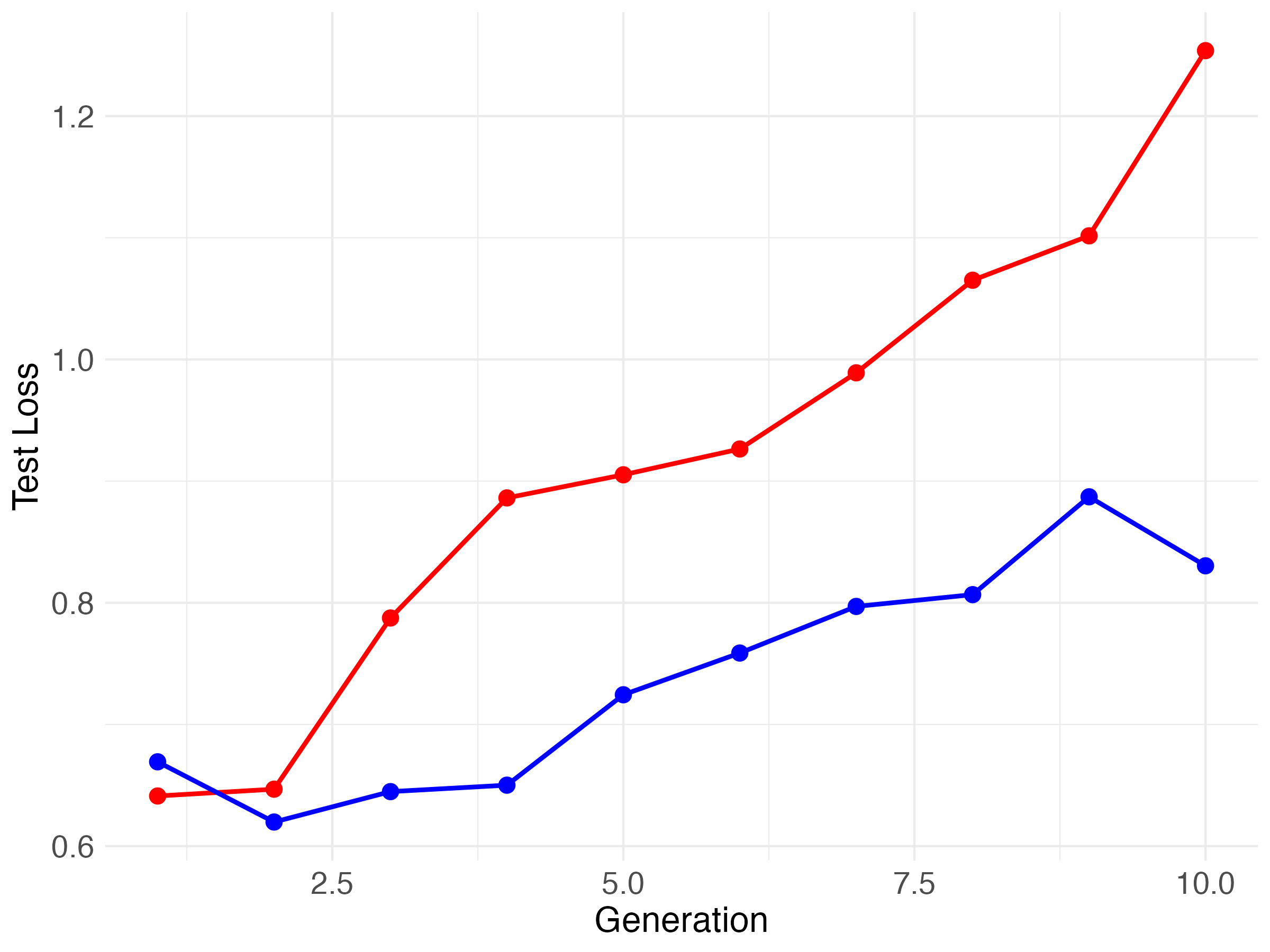}
        \subcaption{Tico}
    \end{subfigure}
    \begin{subfigure}[b]{0.3\textwidth}
        \centering
        \includegraphics[width=\textwidth]{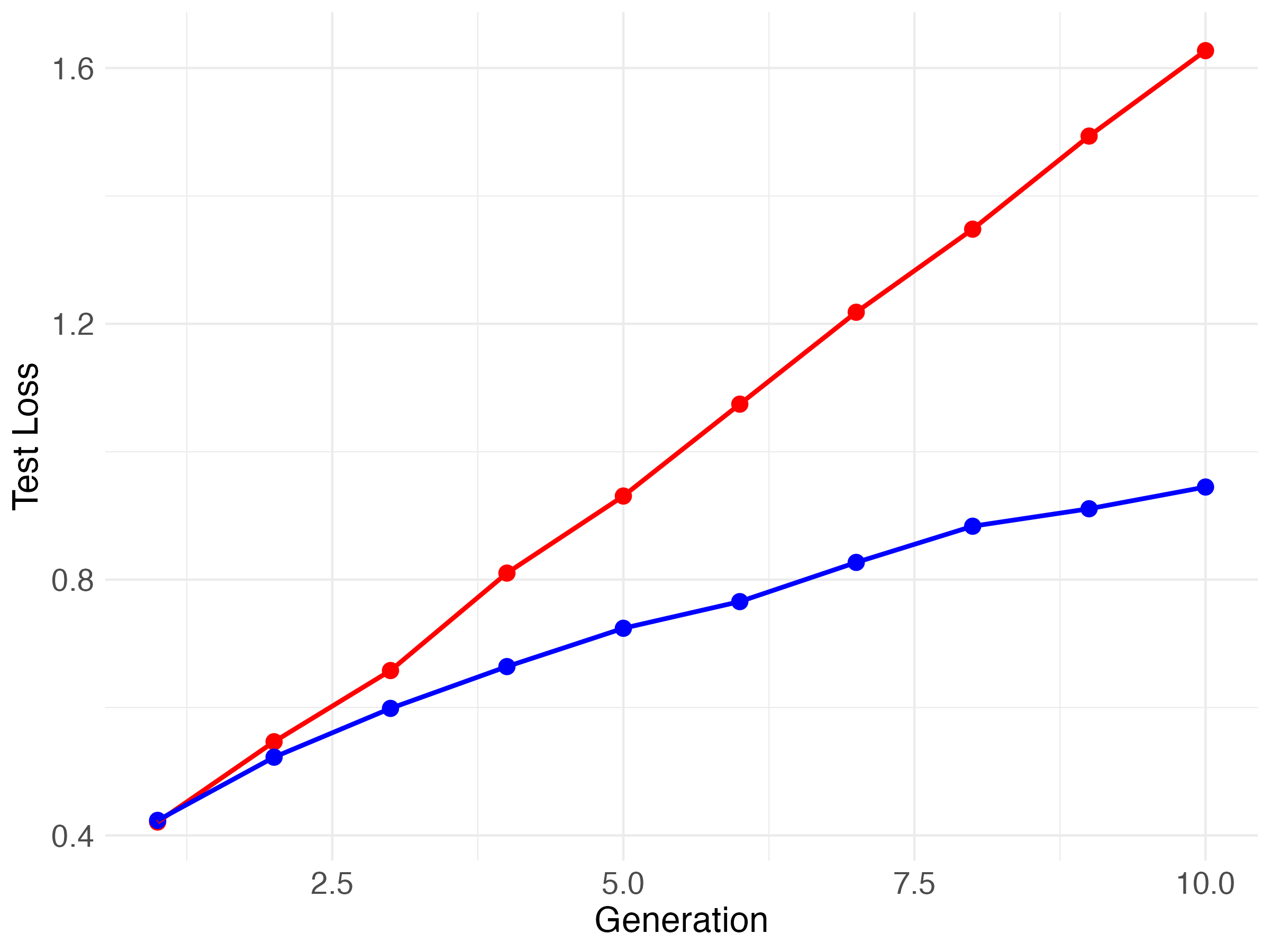}
        \subcaption{VicReg}
    \end{subfigure}

    \caption{Classification test losses for six SSL trained features obtained from ResNet-50 applied on CIFAR-10. \color{red}Red \color{black}denotes \textit{discard} while \color{blue}blue \color{black} denotes \textit{augment}. The blue curves clearly increase at a slower rate.}
    \label{fig:SSL_more}
\end{figure*}

In addition to the experiments presented in Section \ref{sec:experiments} on iterative logistic fit on SSL trained features, we have considered other SSL models as well. The results are shown in Figure \ref{fig:SSL_more}. The SSL checkpoints are taken from \citet{lightly}. For all these models, the curves corresponding to \textit{augment} rise more slowly than those corresponding to \textit{discard}. As mentioned in the main text, these curves have not been obtained from pure logistic fits. The feature dimensionality was too large, preventing us from being able to use tools like \verb|LogisticRegression()| or \verb|glm()|. We actually trained a logistic model with minibatch stochastic gradient descent (SGD) and stopped early. As a result, we do not expect the logistic estimates to have been the most honest ones obtained from convex optimization. As a follow-up, we are performing experiments with subsampled SSL features so that we actually can fit honest logistic models, and results will be reported in the next revision. 

\subsection{Test accuracies for SSL trained iterative logistic fit}

\begin{figure*}[h]
    \centering
    \resizebox{\textwidth}{!}{
        \begin{tabular}{cccc}
            \begin{subfigure}[b]{0.23\textwidth}
                \centering
                \includegraphics[width=\textwidth]{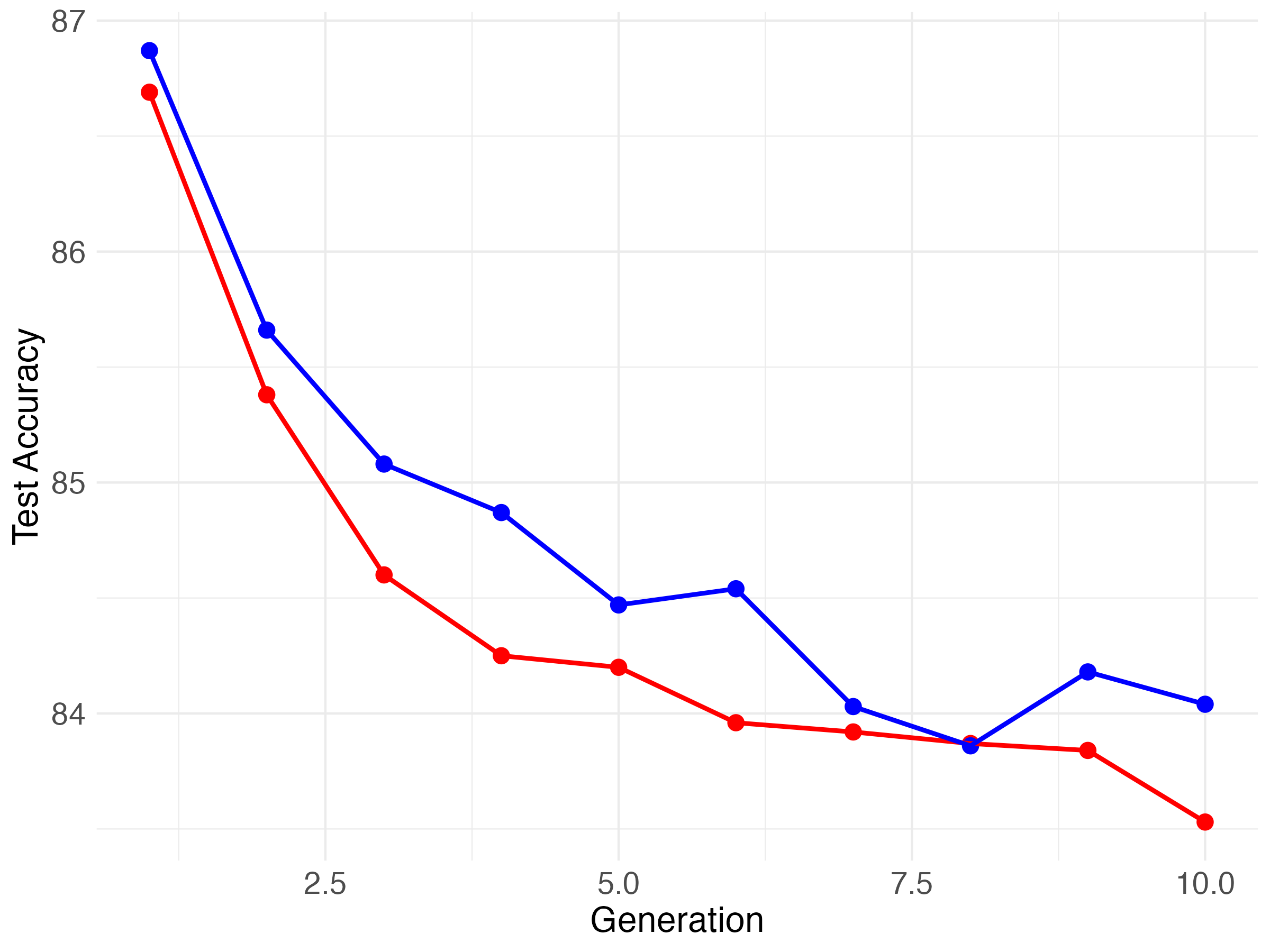}
                \subcaption{SimCLR}
            \end{subfigure} &
            \begin{subfigure}[b]{0.23\textwidth}
                \centering
                \includegraphics[width=\textwidth]{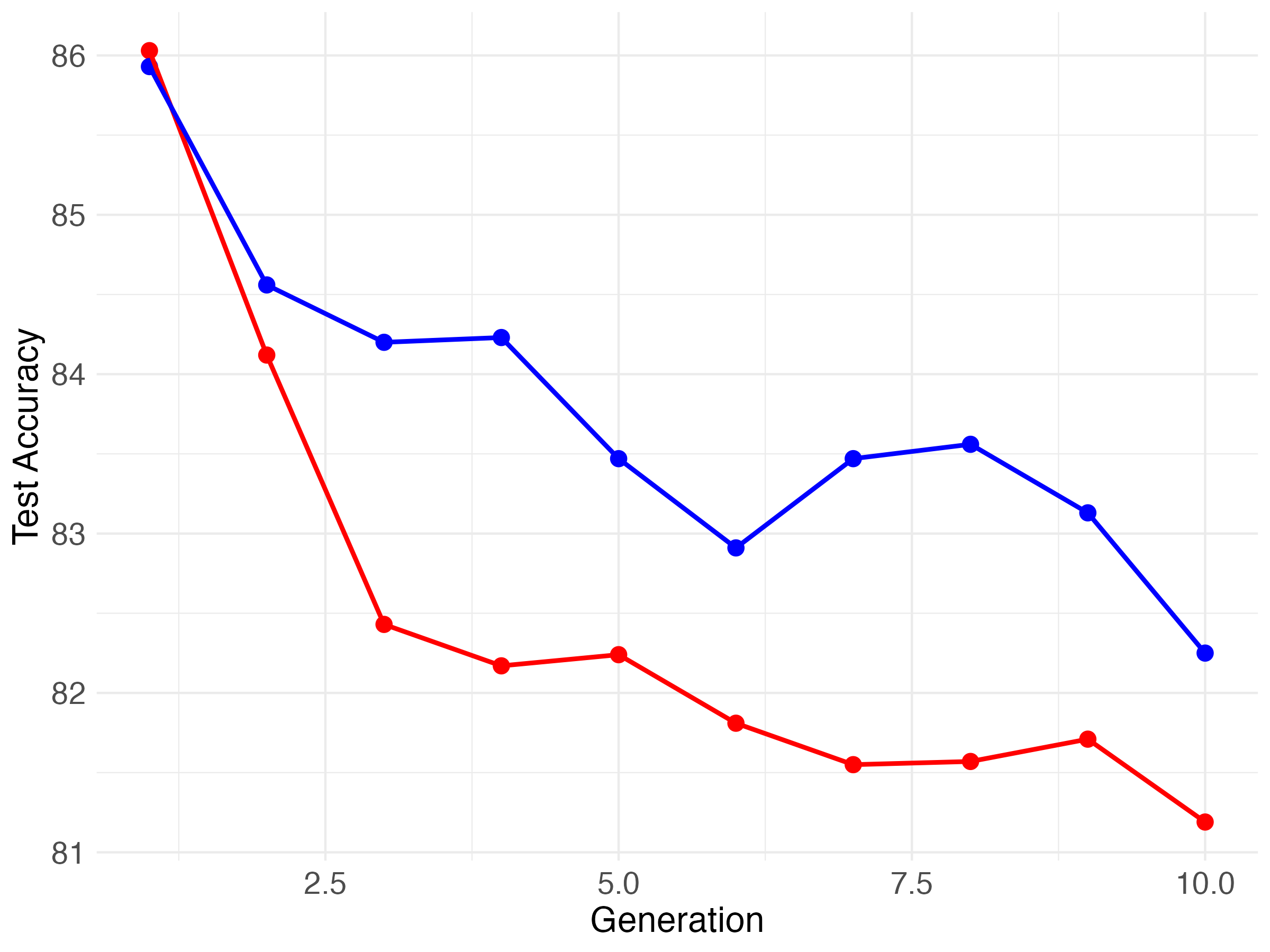}
                \subcaption{DINO}
            \end{subfigure} &
            \begin{subfigure}[b]{0.23\textwidth}
                \centering
                \includegraphics[width=\textwidth]{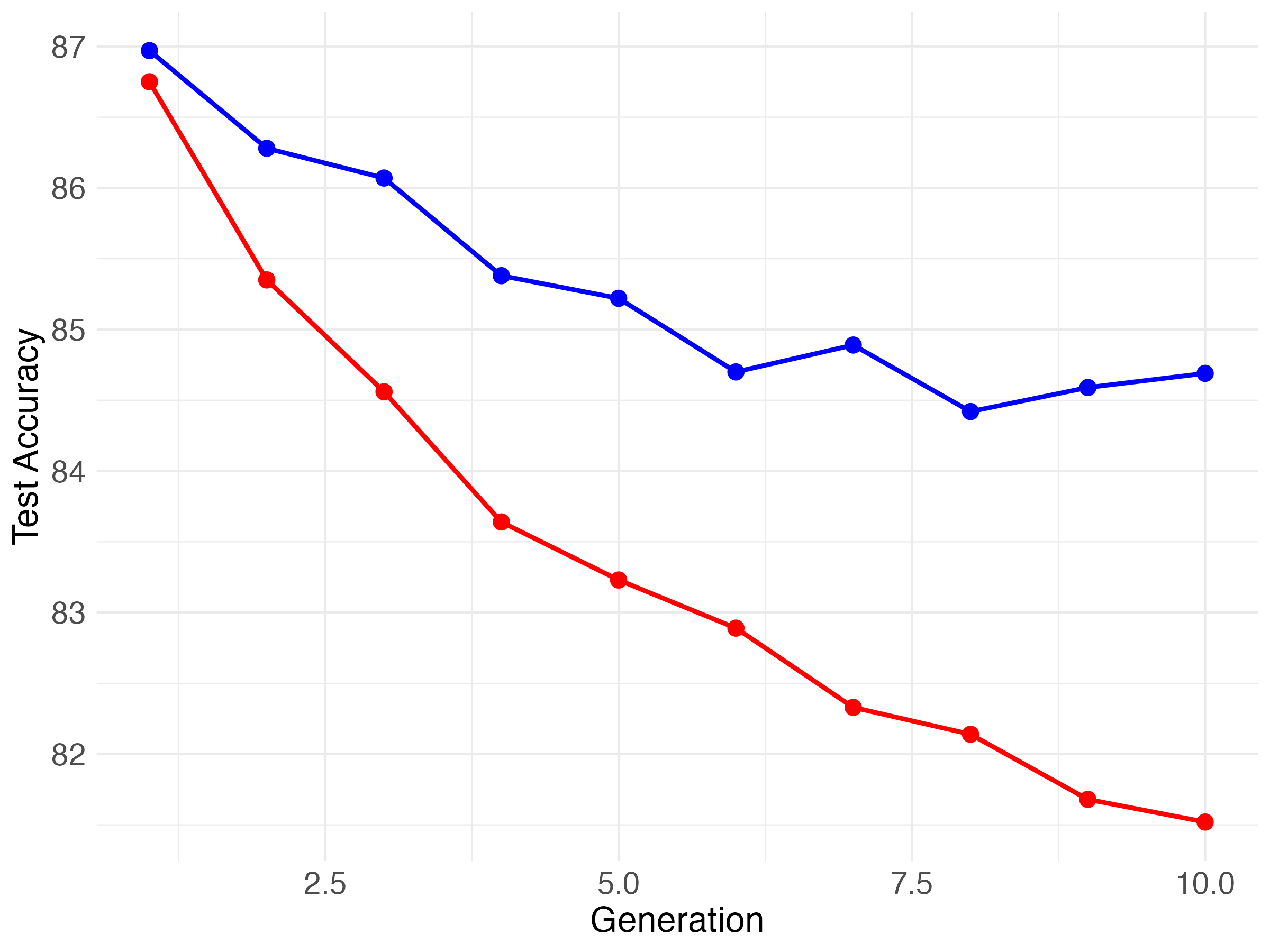}
                \subcaption{MoCoV2}
            \end{subfigure} &
            \begin{subfigure}[b]{0.23\textwidth}
                \centering
                \includegraphics[width=\textwidth]{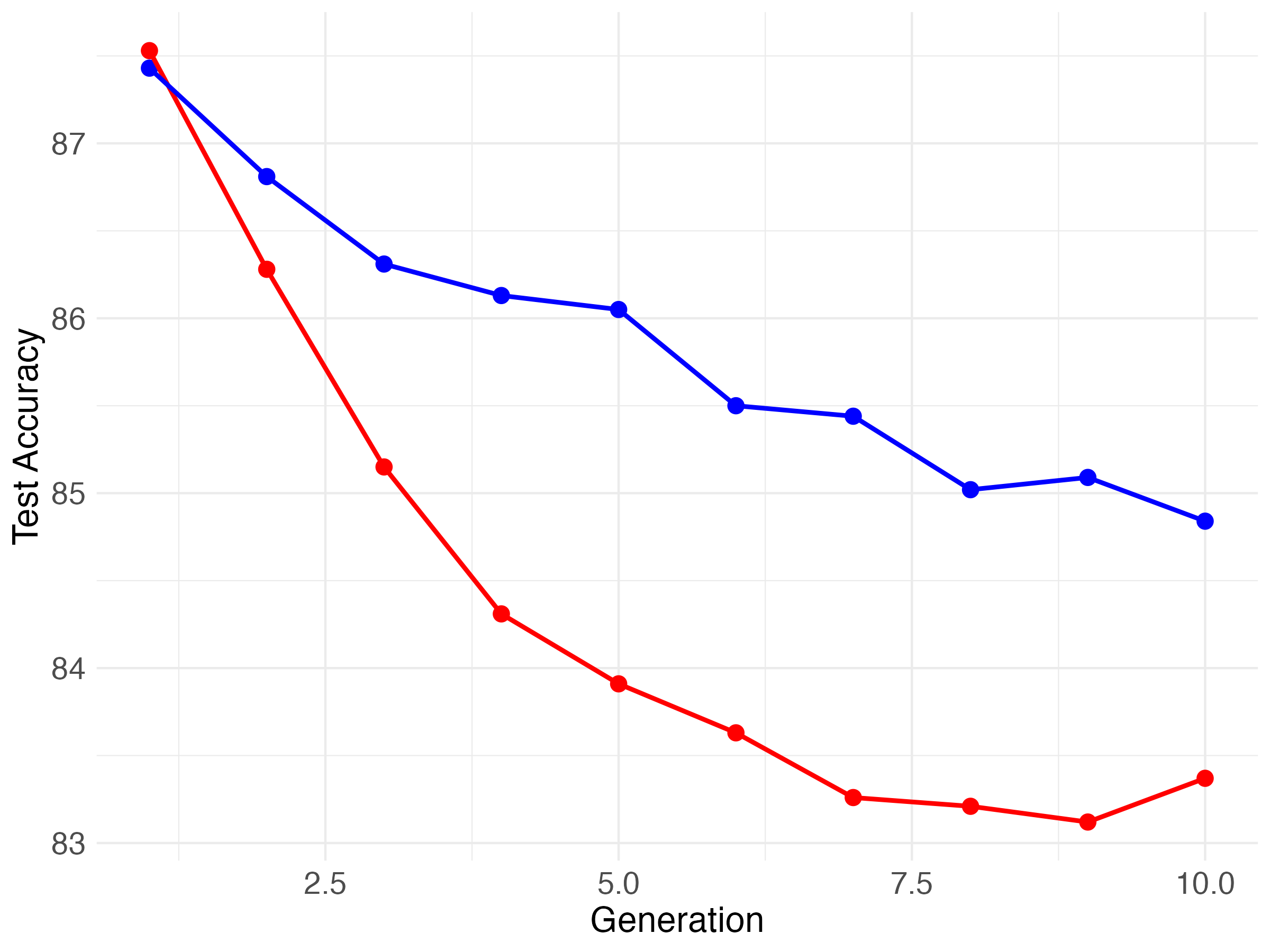}
                \subcaption{SWAV}
            \end{subfigure} \\
            \begin{subfigure}[b]{0.23\textwidth}
                \centering
                \includegraphics[width=\textwidth]{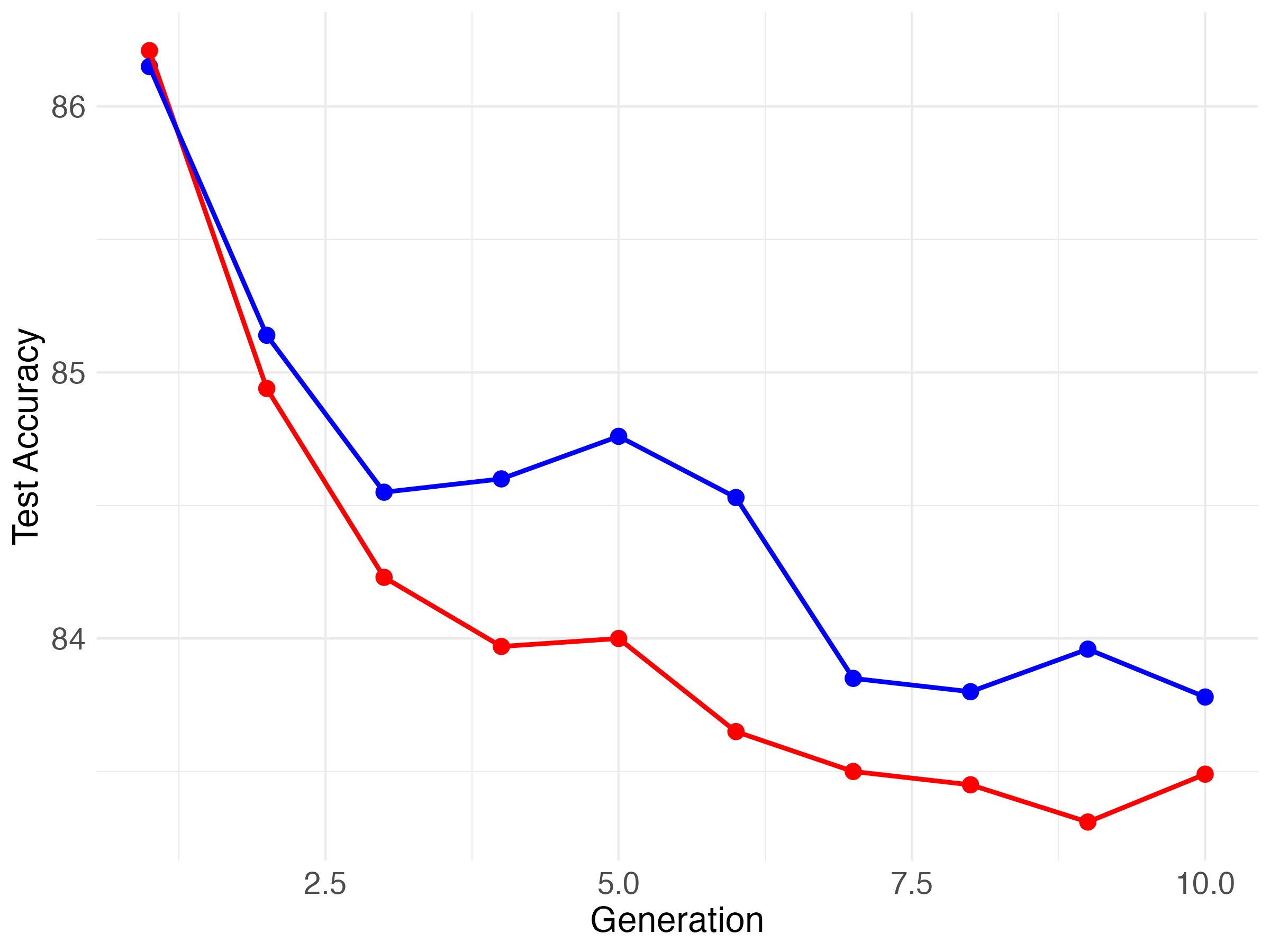}
                \subcaption{Barlow Twins}
            \end{subfigure} &
            \begin{subfigure}[b]{0.23\textwidth}
                \centering
                \includegraphics[width=\textwidth]{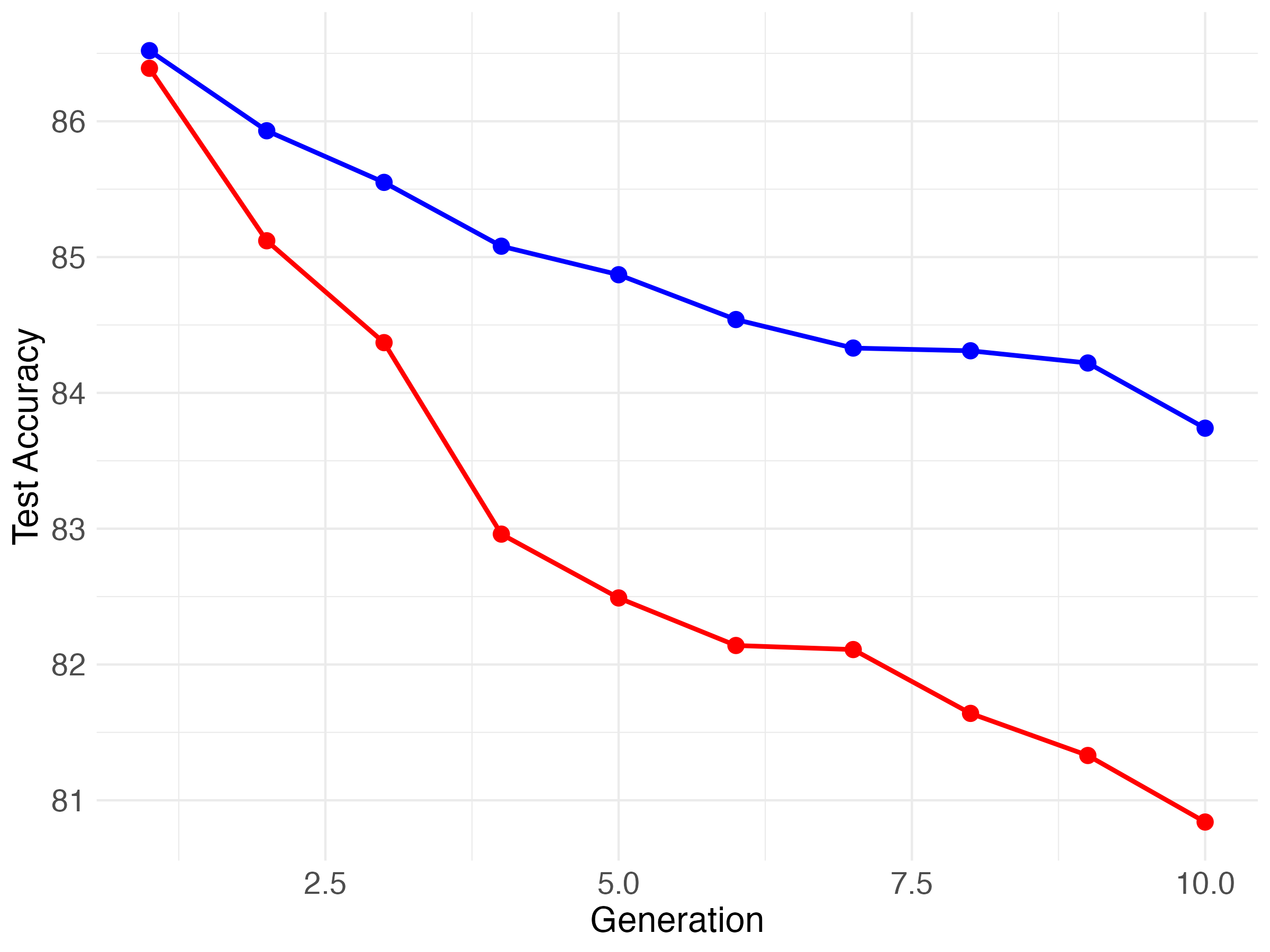}
                \subcaption{Byol}
            \end{subfigure} &
            \begin{subfigure}[b]{0.23\textwidth}
                \centering
                \includegraphics[width=\textwidth]{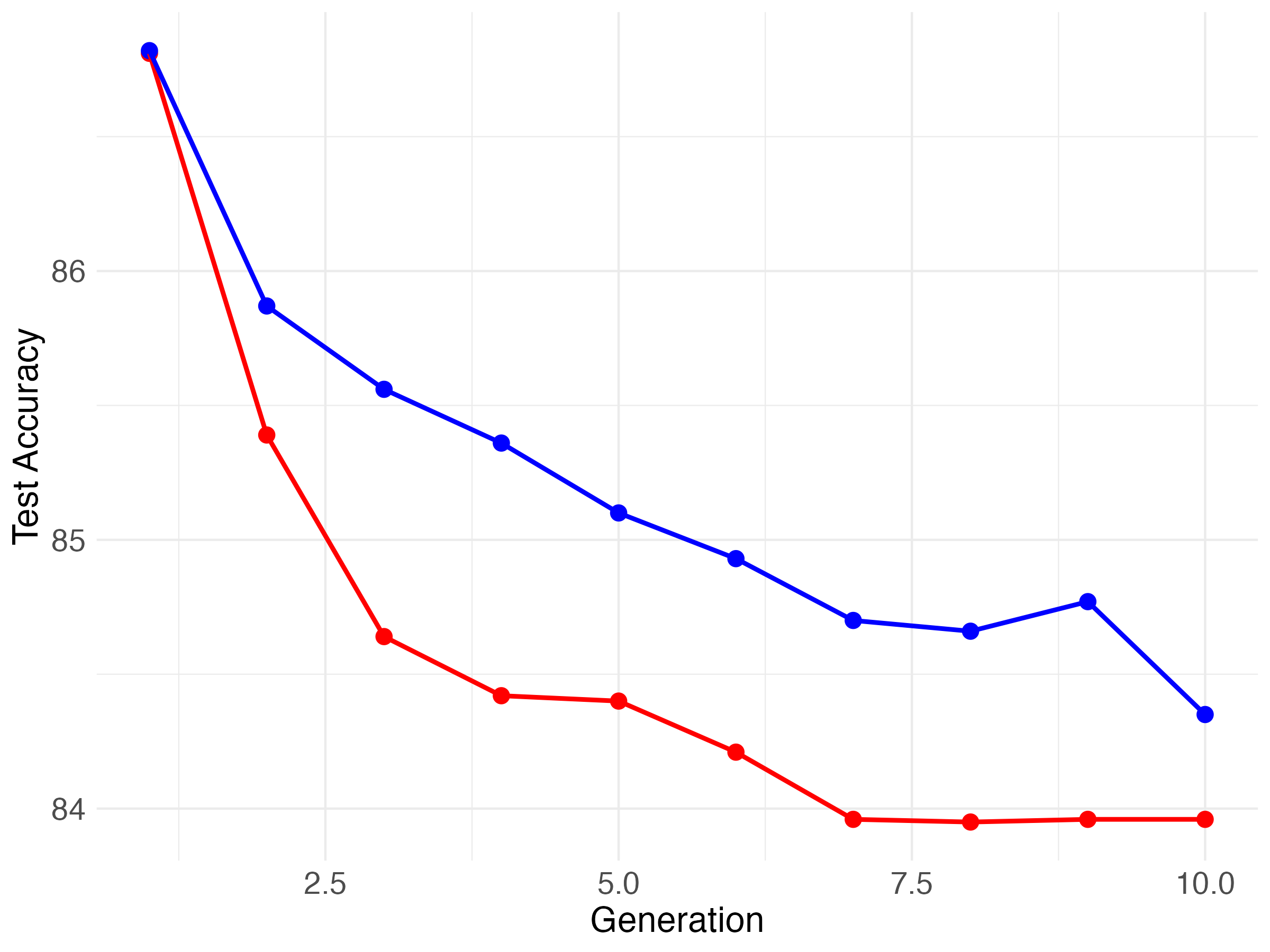}
                \subcaption{DCL}
            \end{subfigure} &
            \begin{subfigure}[b]{0.23\textwidth}
                \centering
                \includegraphics[width=\textwidth]{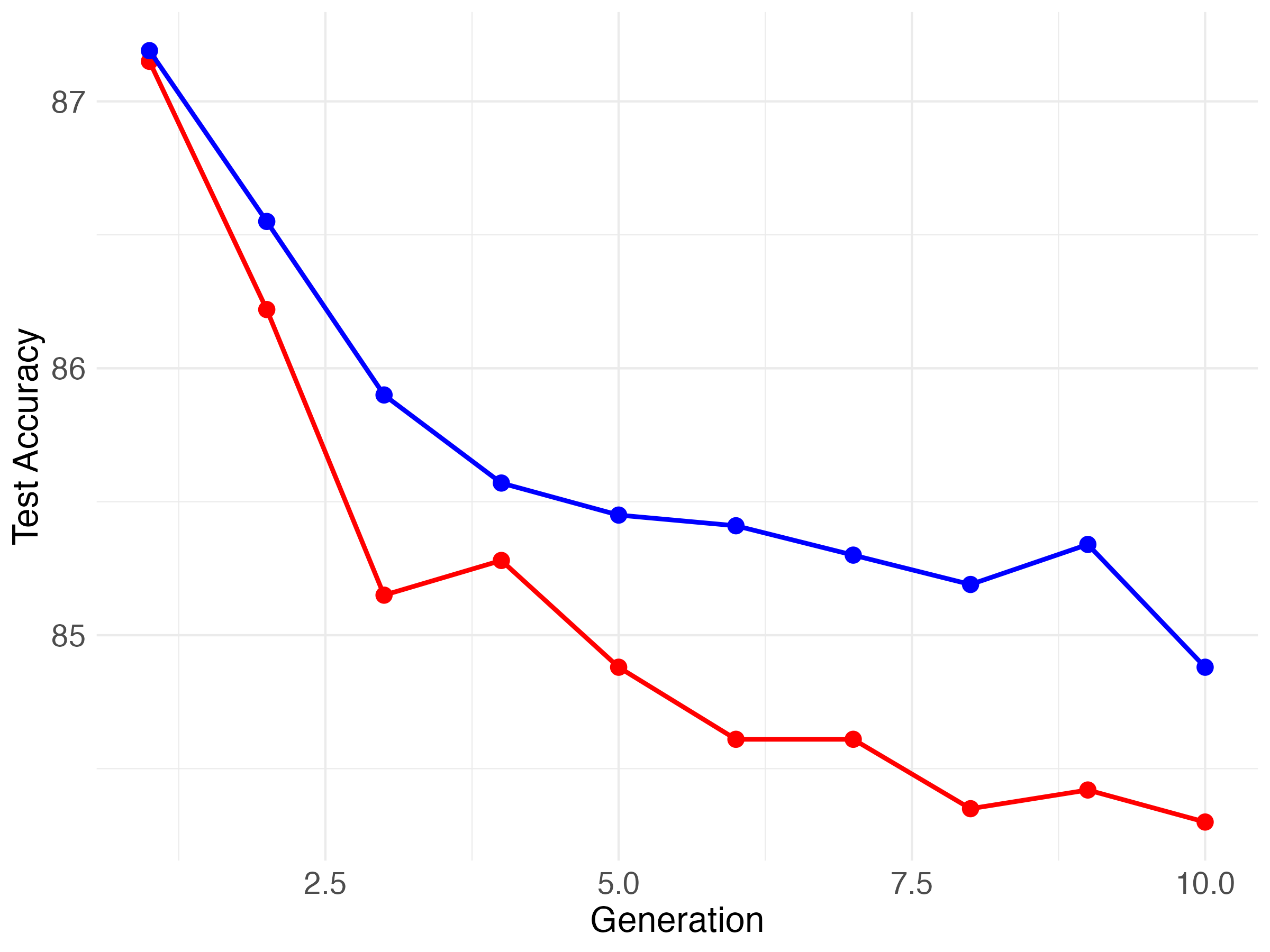}
                \subcaption{DCLW}
            \end{subfigure} \\
            \begin{subfigure}[b]{0.23\textwidth}
                \centering
                \includegraphics[width=\textwidth]{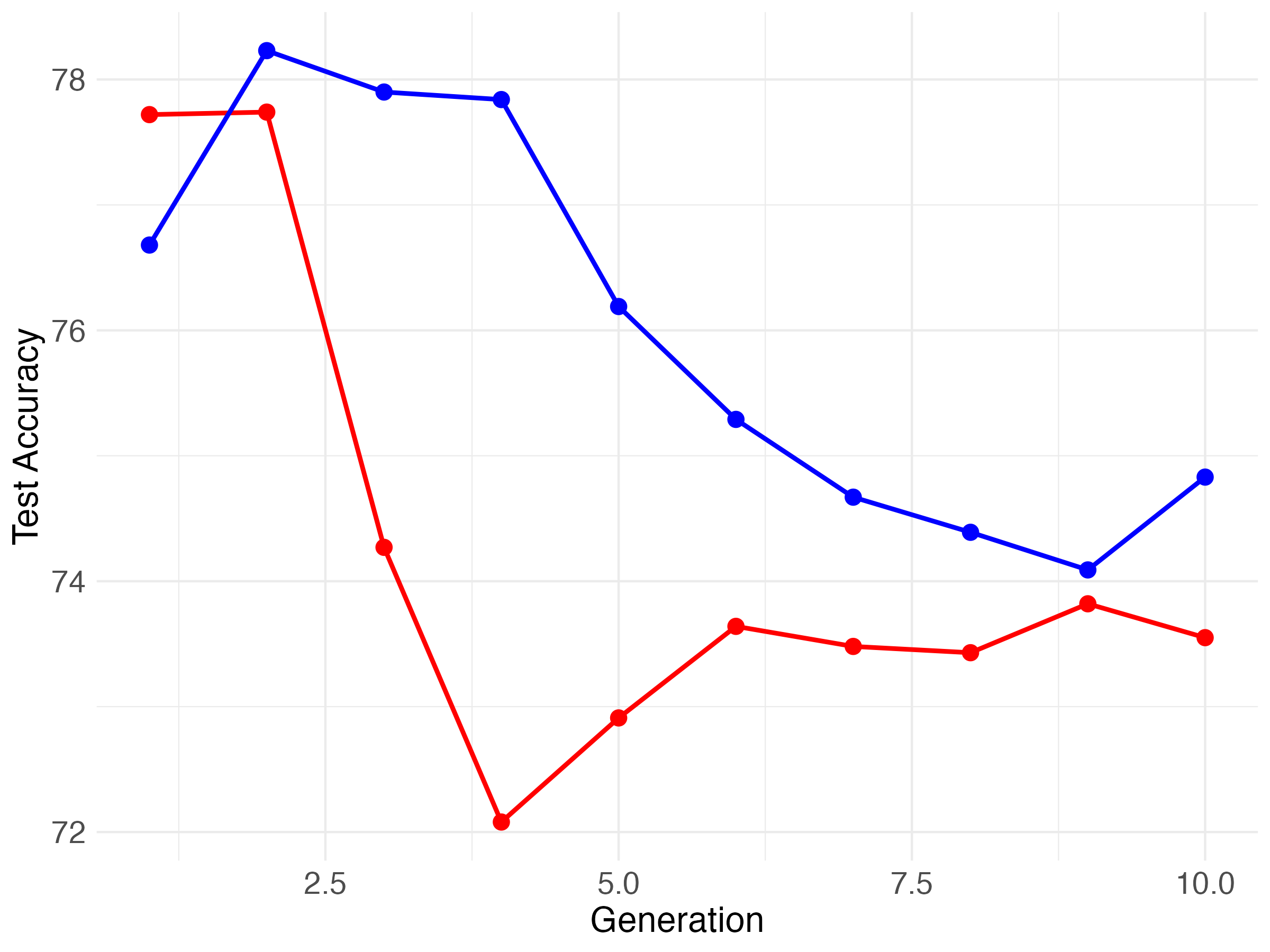}
                \subcaption{Tico}
            \end{subfigure} &
            \begin{subfigure}[b]{0.23\textwidth}
                \centering
                \includegraphics[width=\textwidth]{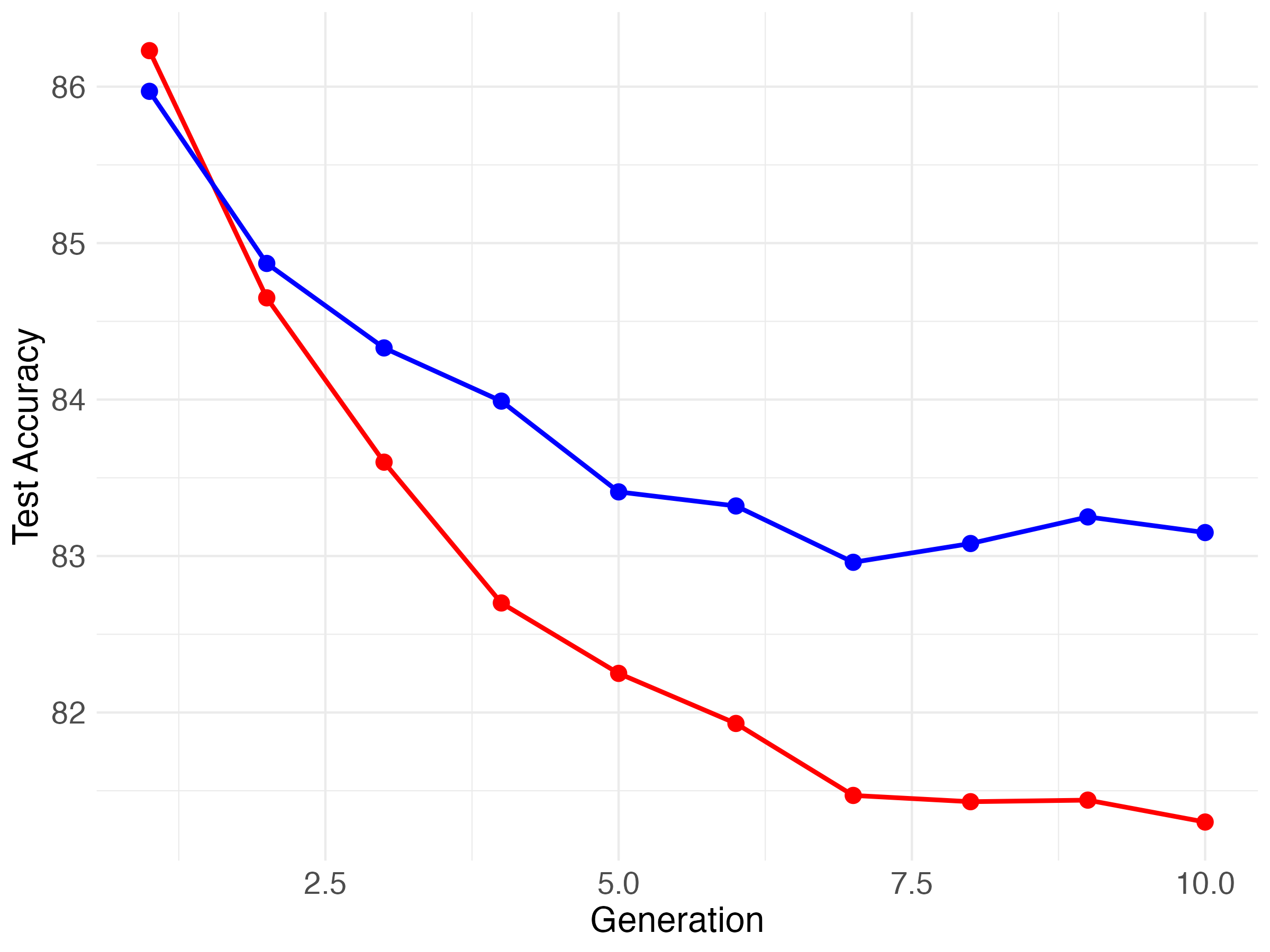}
                \subcaption{VicReg}
            \end{subfigure}
        \end{tabular}
    }
    \caption{Classification test accuracies for ten SSL trained features obtained from ResNet-50 applied on CIFAR-10. \color{red}Red \color{black}denotes \textit{discard} while \color{blue}blue \color{black} denotes \textit{augment}. The blue curves clearly decrease at a slower rate.}
    \label{fig:SSL_accuracies}
\end{figure*}

We now provide the test accuracy curves on the iterative logistic fits obtained from the 10 SSL models, in Figure \ref{fig:SSL_accuracies}.

\end{document}